\documentclass[letterpaper, 10 pt, conference]{ieeeconf}  %

\IEEEoverridecommandlockouts                              %

\usepackage{graphicx}

\usepackage{tikz}
\usepackage{comment}
\usepackage{amsmath,amssymb} %

\usepackage[nocompress]{cite}

\usepackage{etoolbox}

\usepackage{graphicx}
\usepackage{comment}
\usepackage{amsmath,amssymb} %
\usepackage{balance}
\usepackage{flushend}

\usepackage{booktabs}

\usepackage{wrapfig}

\usepackage{longtable}
\usepackage{epstopdf}
\usepackage[permil]{overpic}
\usepackage{arydshln}
\usepackage{multirow}

\usepackage{xcolor}
\usepackage{colortbl}

\usepackage{microtype} 
\usepackage{xspace}

\usepackage{pbox}
\usepackage{pifont}
\usepackage{array} %
\let\labelindent\relax

\usepackage{enumitem}

\def\eg{\emph{e.g.}\@\xspace} 

\def\ie{\emph{i.e.}\@\xspace}

\def\etal{\emph{et al.}\@\xspace}

\newcommand{\iou}{IoU}
\newcommand{\gt}{ground-truth}

\newcolumntype{s}{>{\columncolor{white!90!blue}} c}
\newcolumntype{g}{>{\columncolor{black!10}} c}

\newcommand{\para}[1]{\vskip2pt \noindent \textbf{#1}}

\definecolor{predictionpurple}{RGB}{204, 186, 240}

\title{\LARGE \bf
Interactive Object Segmentation in 3D Point Clouds
}

\author{Theodora Kontogianni$^{1}$, Ekin Celikkan$^{2}$, Siyu Tang$^{1}$, Konrad Schindler$^{1}$%
\thanks{$^{1}$ETH Z\"{u}rich, Switzerland}%
\thanks{$^{2}$Computer Vision Group, RWTH Aachen University, Germany}%
}

\begin{document}

\definecolor{linkgreen}{RGB}{52,130,48}
\definecolor{darkred}{RGB}{203,65,84}

\maketitle
\thispagestyle{empty}
\pagestyle{empty}

\begin{abstract}
We propose an interactive approach for 3D instance segmentation, where users can iteratively collaborate with a deep learning model to  segment objects in a 3D point cloud directly.
Current methods for 3D instance segmentation are generally trained in a fully-supervised fashion, which requires large amounts of costly training labels, and does not generalize well to classes unseen during training.
Few works have attempted to obtain 3D segmentation masks using human interactions. Existing methods rely on user feedback in the 2D image domain.
As a consequence, users are required to constantly switch between 2D images and 3D representations, and custom architectures are employed to combine multiple input modalities. Therefore, integration with existing standard 3D models is not straightforward.
The core idea of this work is to enable users to  interact directly with 3D point clouds
by clicking on desired 3D objects of interest~(or their background) to interactively segment the scene in an open-world setting. Specifically, our method does not require training data from any target domain, and can adapt to new environments where no appropriate training sets are available.
Our system continuously adjusts the object segmentation based on the user feedback and achieves accurate dense 3D segmentation masks with minimal human effort (few clicks per object). Besides its potential for efficient labeling of large-scale and varied 3D datasets, our approach, where the user directly interacts with the 3D environment, enables new applications in AR/VR and human-robot interaction.

\end{abstract}

\section{Introduction}
\vspace{-5px}

Just like other occurrences of supervised deep learning, 3D scene understanding tasks such as 3D semantic instance segmentation require large amounts of annotated training data.
Alas, in 3D, dense manual annotation is an even more time-consuming and tedious effort.
Interactive object segmentation techniques are commonly used in the 2D image domain to collect ground-truth segmentation masks at scale\,\cite{benenson19cvpr} or on image editing applications. However, interactive object segmentation in the 3D domain is largely under-explored. 

\begin{figure}[t]
\centering
\setlength{\tabcolsep}{1pt}

\begin{tabular}{cccc}

\includegraphics[trim={5cm 0 5cm 0cm},clip, height=1.7cm]{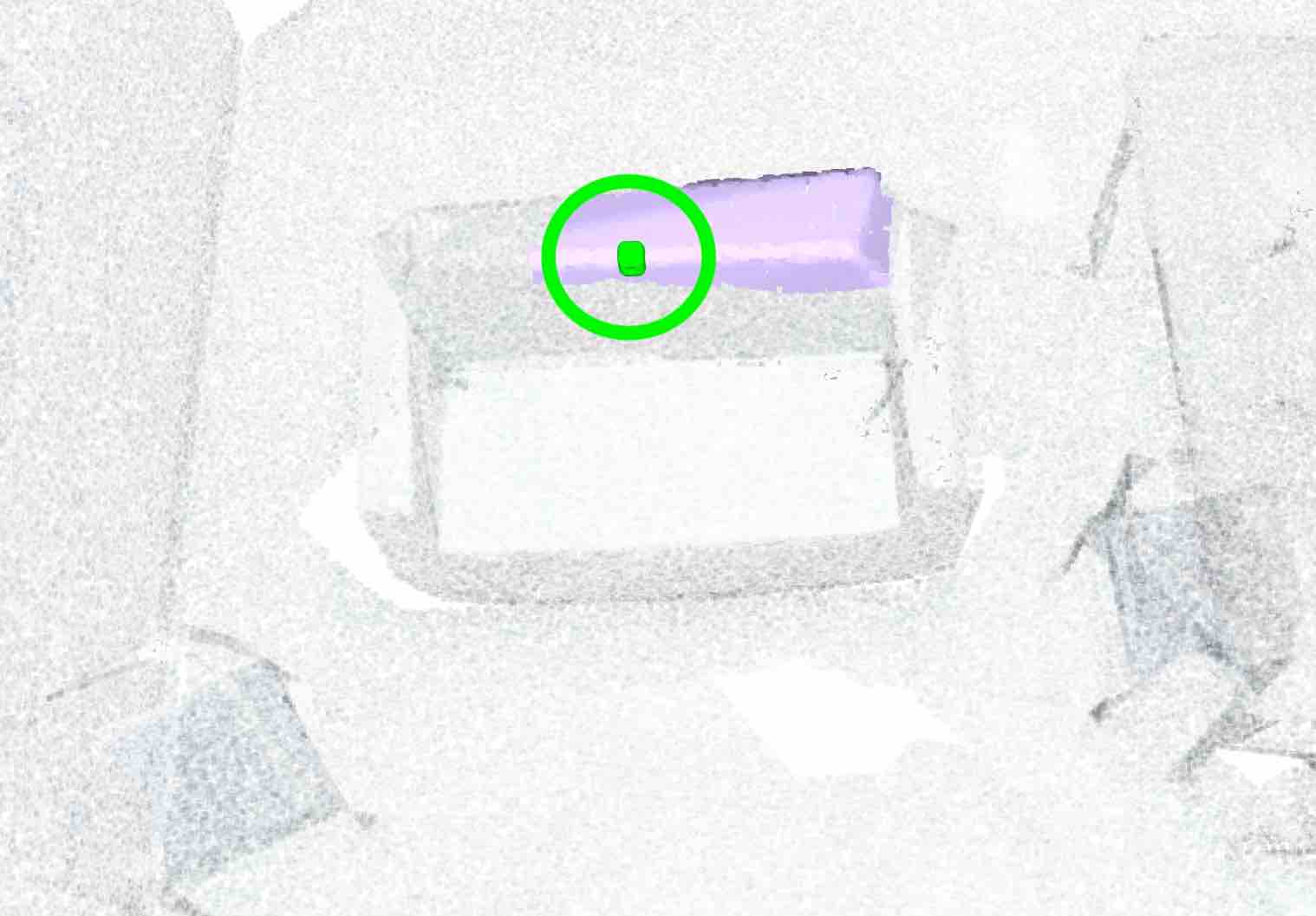} & \includegraphics[trim={5cm 0 5cm 0cm},clip, height=1.7cm]{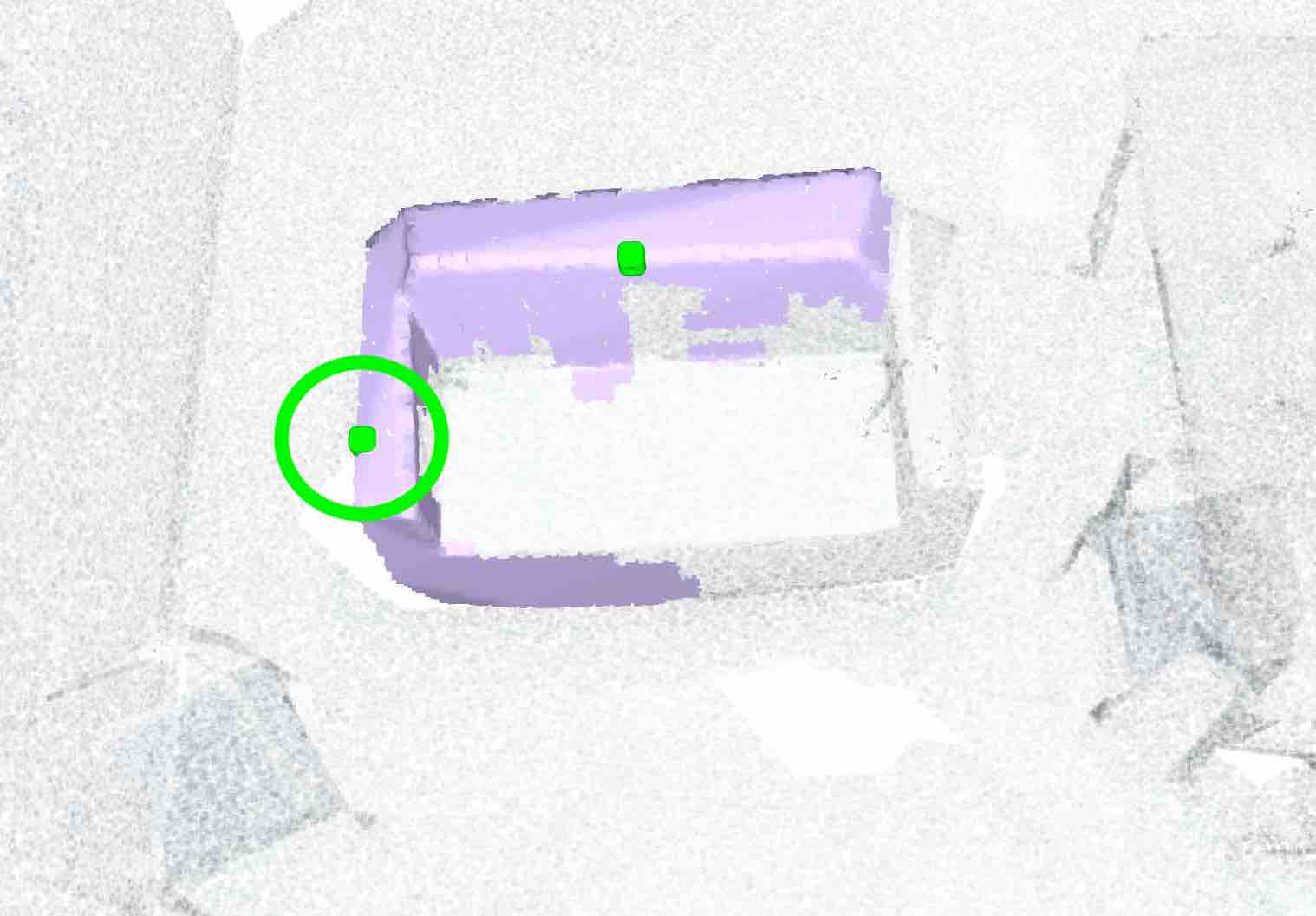} & 
\includegraphics[trim={5cm 0 5cm 0cm},clip, height=1.7cm]{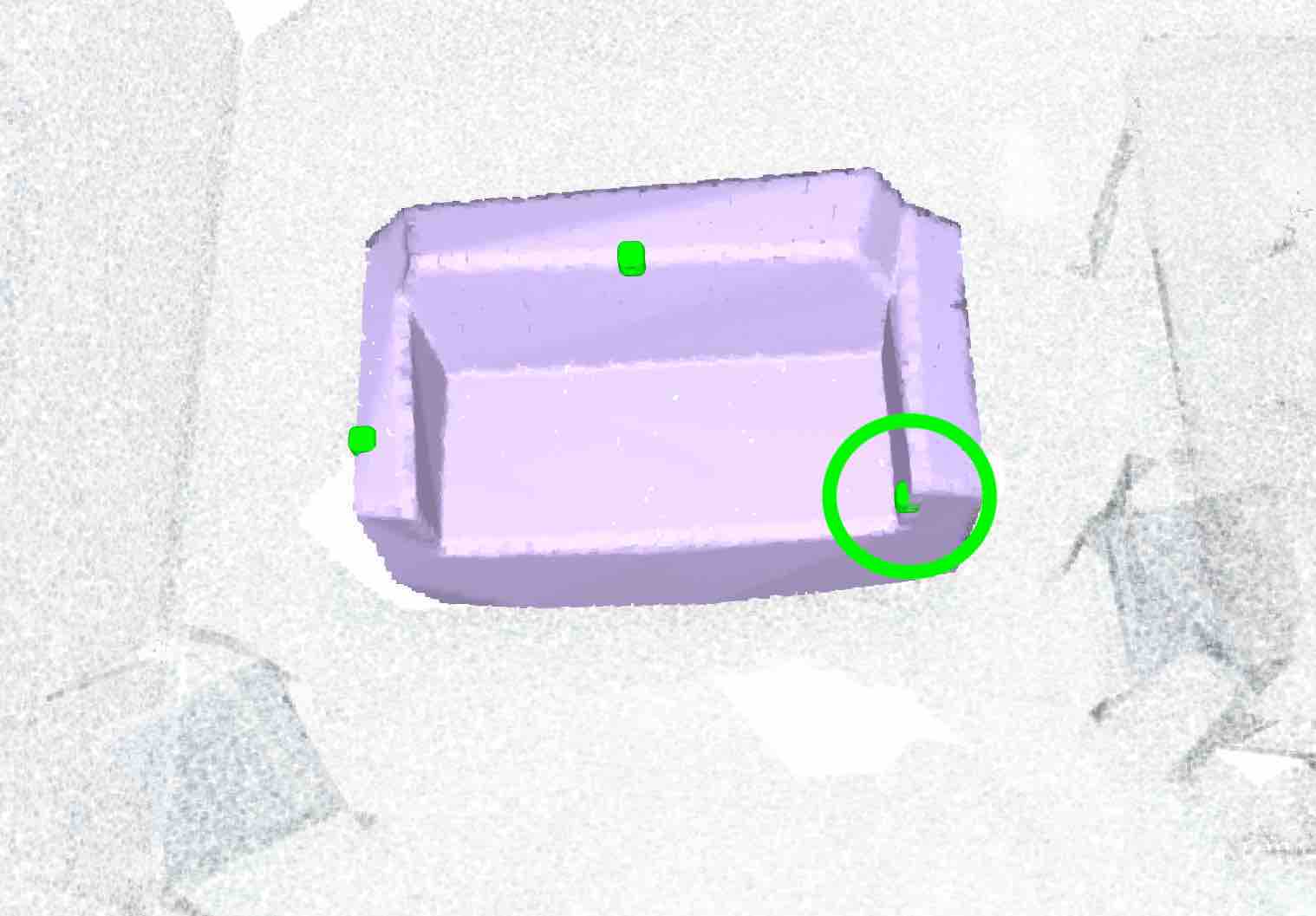} &
\includegraphics[trim={5cm 0 5cm 0cm},clip, height=1.7cm]{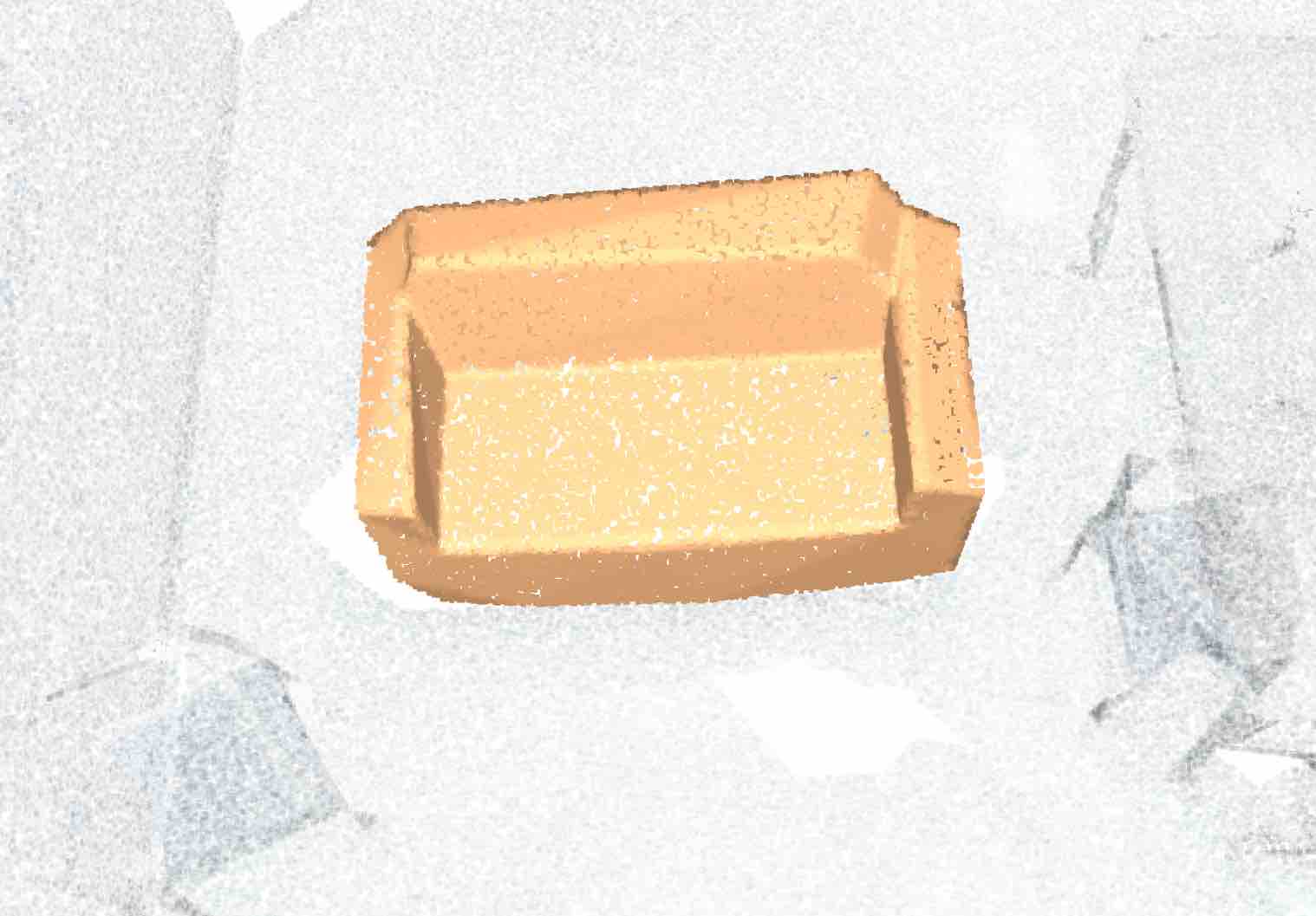}\\

\includegraphics[trim={5cm 5cm 5cm 5cm},clip, height=1.9cm]{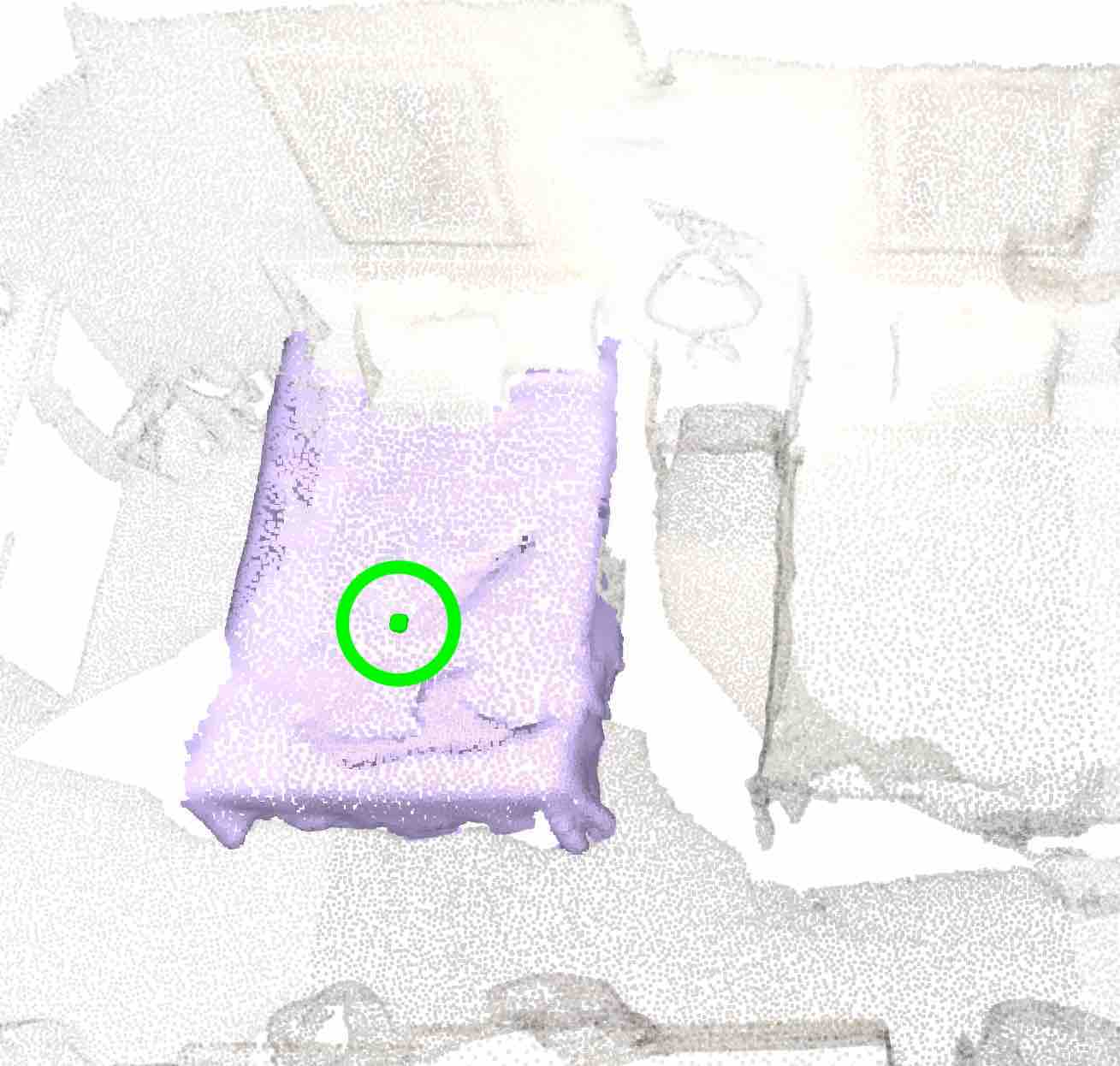}
& \includegraphics[trim={5cm 5cm 5cm 5cm},clip, height=1.9cm]{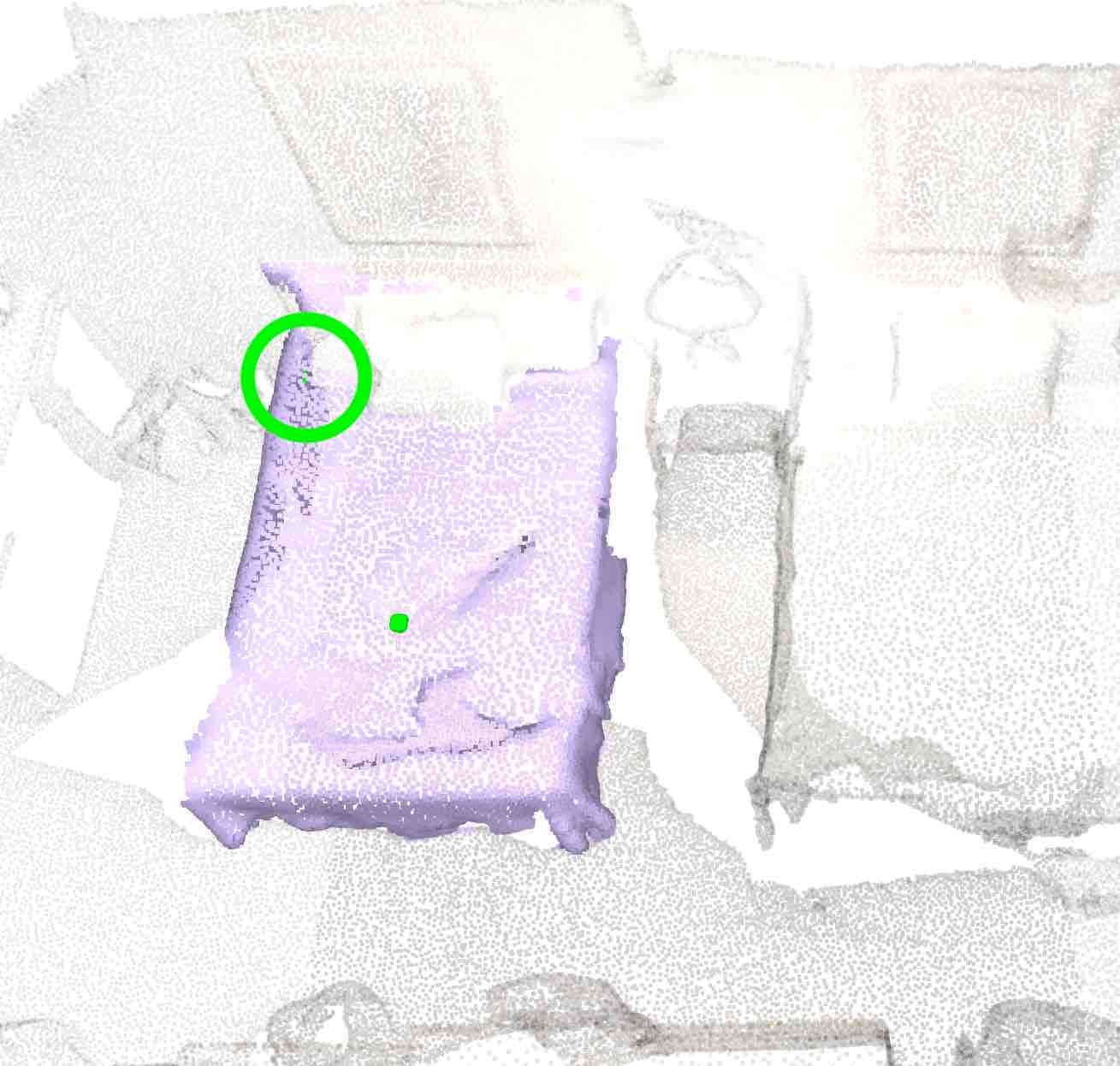} & 
\includegraphics[trim={5cm 5cm 5cm 5cm},clip, height=1.9cm]{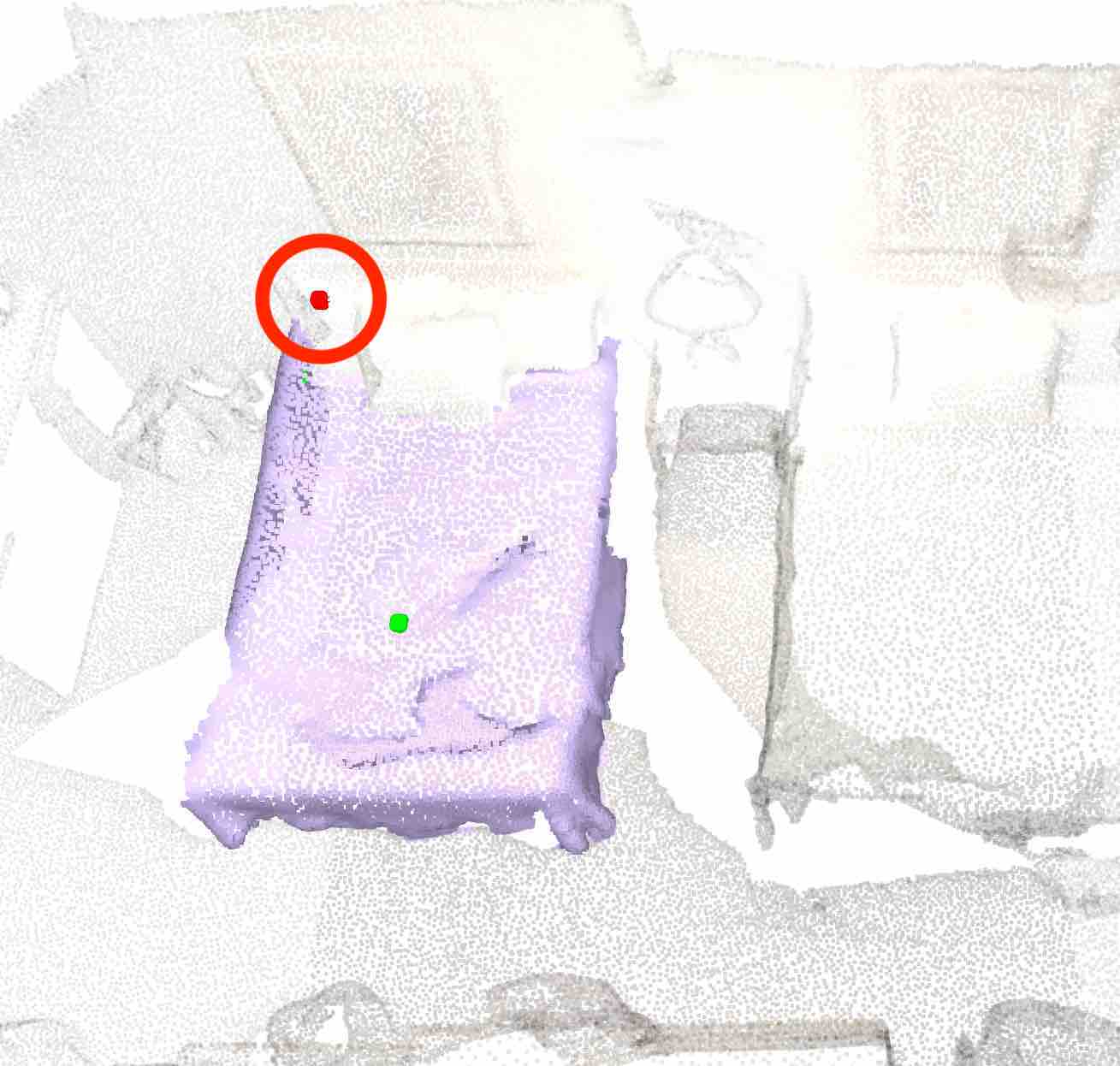} &
\includegraphics[trim={5cm 5cm 5cm 5cm},clip, height=1.9cm]{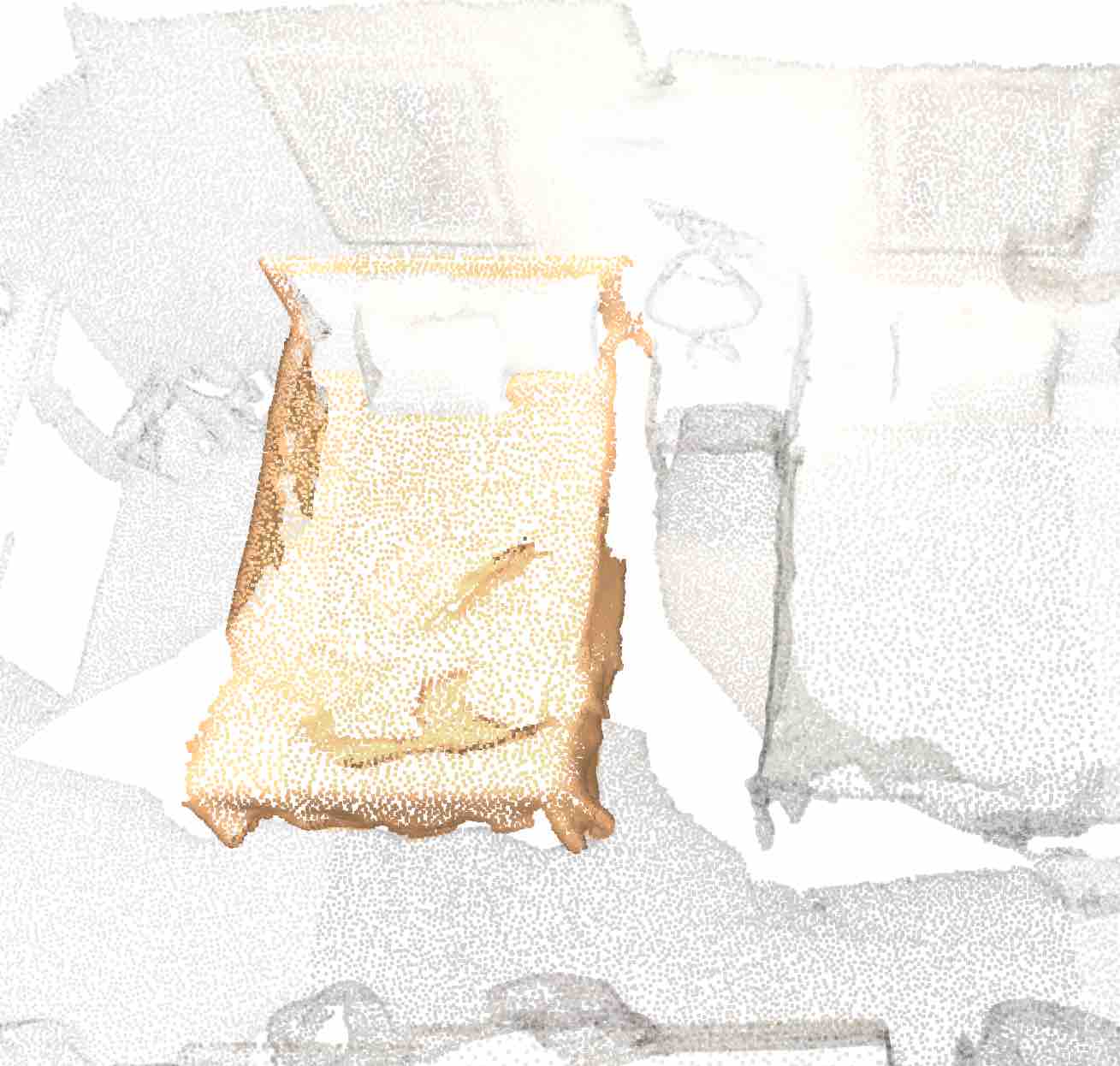}\\
1 click &2 clicks & 3 clicks& Ground Truth\\
\end{tabular}

\caption{\textbf{Interactive 3D Object Segmentation.}
The model predicts a segmentation mask and iteratively updates it, driven by user feedback in the form of \textbf{\textcolor{linkgreen}{positive}} clicks on false negatives, and \textbf{\textcolor{darkred}{negative}} clicks on false positives.
The current click is visualized as enlarged circle.
}
\label{fig:inter_obj_example}
\end{figure}

The principle of interactive object segmentation \cite{xu16cvpr,liew17iccv,mahadevan18bmvc,li18cvpr,benenson19cvpr,jang19cvpr} is to let a user collaborate with a model when segmenting an object instance.
By concentrating on one instance  at a time (\eg, a specific chair), the task becomes a binary foreground/background segmentation.
The segmentation mask is obtained with an iterative
procedure: the model produces its best guess of a mask, then the user provides feedback via corrections, which the model ingests to produce an updated mask.
That sequence is repeated until the mask is deemed accurate enough by the user (Fig.\ref{fig:inter_obj_example}).
The corrections usually come in the form of clicks that identify locations with incorrect labels. \textit{Negative} clicks identify false positives, where parts of the background are labeled as foreground. \textit{Positive} clicks identify false negatives where parts of the object have been falsely labeled as background.

\begin{figure}[!t]
\centering
\begin{small}
\begin{tabular}{cccc}
    \includegraphics[trim={0 0 0 0cm},clip, height=1.5cm]{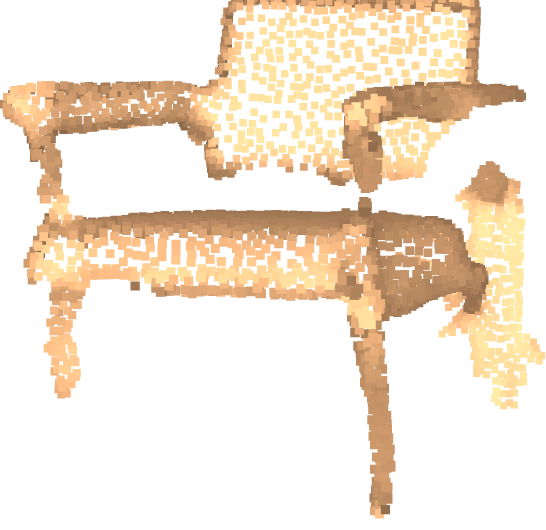} &     \hfill\includegraphics[trim={0 0 0 0cm},clip, height=1.5cm]{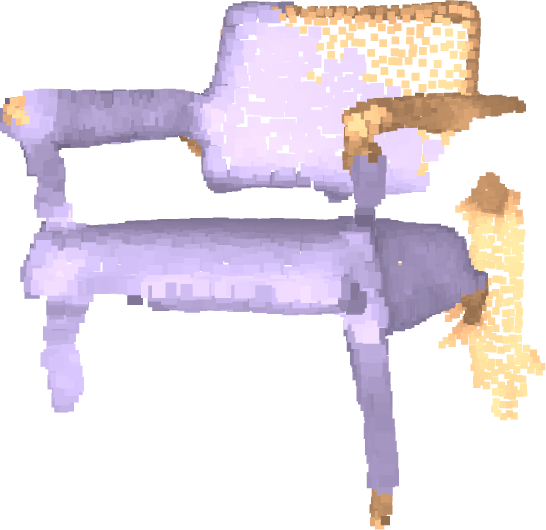} & 
    \includegraphics[trim={0 0 0 0cm},clip, height=1.5cm]{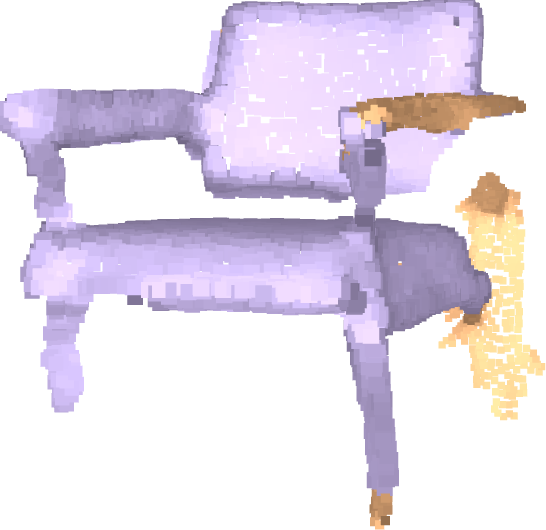}&
    \includegraphics[trim={0 0 0 0cm},clip, height=1.5cm]{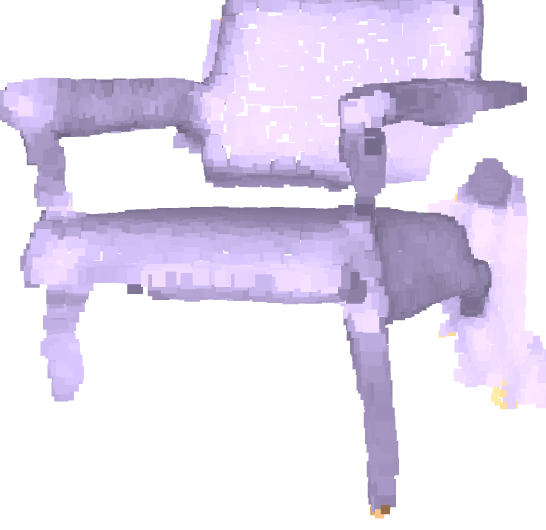}\\
Ground Truth & \iou:\,60\% & \iou:\,80\% &\iou:\,90\% \\
\end{tabular}
\end{small}
\vspace{-3px}
\caption{
We believe that high quality segmentation masks, that can for example be used as ground truth, should be at least above 90\,\%\,\iou{}.
This performance cannot yet be reached, even with fully-supervised instance segmentation methods~(Tab.~\ref{tab:comparison_w_instance}).
As shown above, a segmentation mask (\color{predictionpurple}{{purple}}\color{black}{)} with 80\%\,\iou{} still has significant erroneous regions.
This motivates our approach for \emph{interactive} 3D instance segmentation, where a user can significantly increase the segmentation quality, with just a few clicks.}
\label{fig:iou_level}
\end{figure}

Even though interactive segmentation in images is a well-established technology, and part of several consumer software products, little is known about its potential for labeling 3D point clouds. Despite the recent progress of 3D deep learning, and the persistent dearth of realistic large-scale training sets,  interactive 3D object segmentation remains under-explored.
 
The field of 3D deep learning has witnessed significant progress in the past few years, particularly regarding semantic and instance segmentation in 3D point clouds~\cite{jiang2021iccv,wang2020arxiv,zhang2021iccv,liu2021cvpr,hou2021cvpr,xu2020cvpr}.
Nonetheless, existing instance segmentation methods are not suitable for labeling new datasets.
They require lots of training data to begin with.
More importantly, they cannot generalize to classes that were not part of the training set, a situation commonly encountered when labeling new datasets.
What is more, the resulting segmentation masks tend to have high, but imperfect accuracy (\textless\,90\%\,IoU), which is impressive for an AI system, but not enough to serve as high-quality ground truth.
We argue that even segmentation masks of \,80\%\,IoU are sub-optimal in representing a 3D object accurately~(see Fig.~\ref{fig:iou_level}). These masks are too inaccurate to represent ground truth. Note, though, that our proposed interactive method can be integrated in most existing instance segmentation algorithms to obtain improved segmentation masks with little user input.

 \begin{figure*}[!t]
 \vspace{5px}

\begin{small}
\begin{tabular}{ccc}
    \includegraphics[height=0.16\textwidth, trim={0 250px 0 0}, clip]{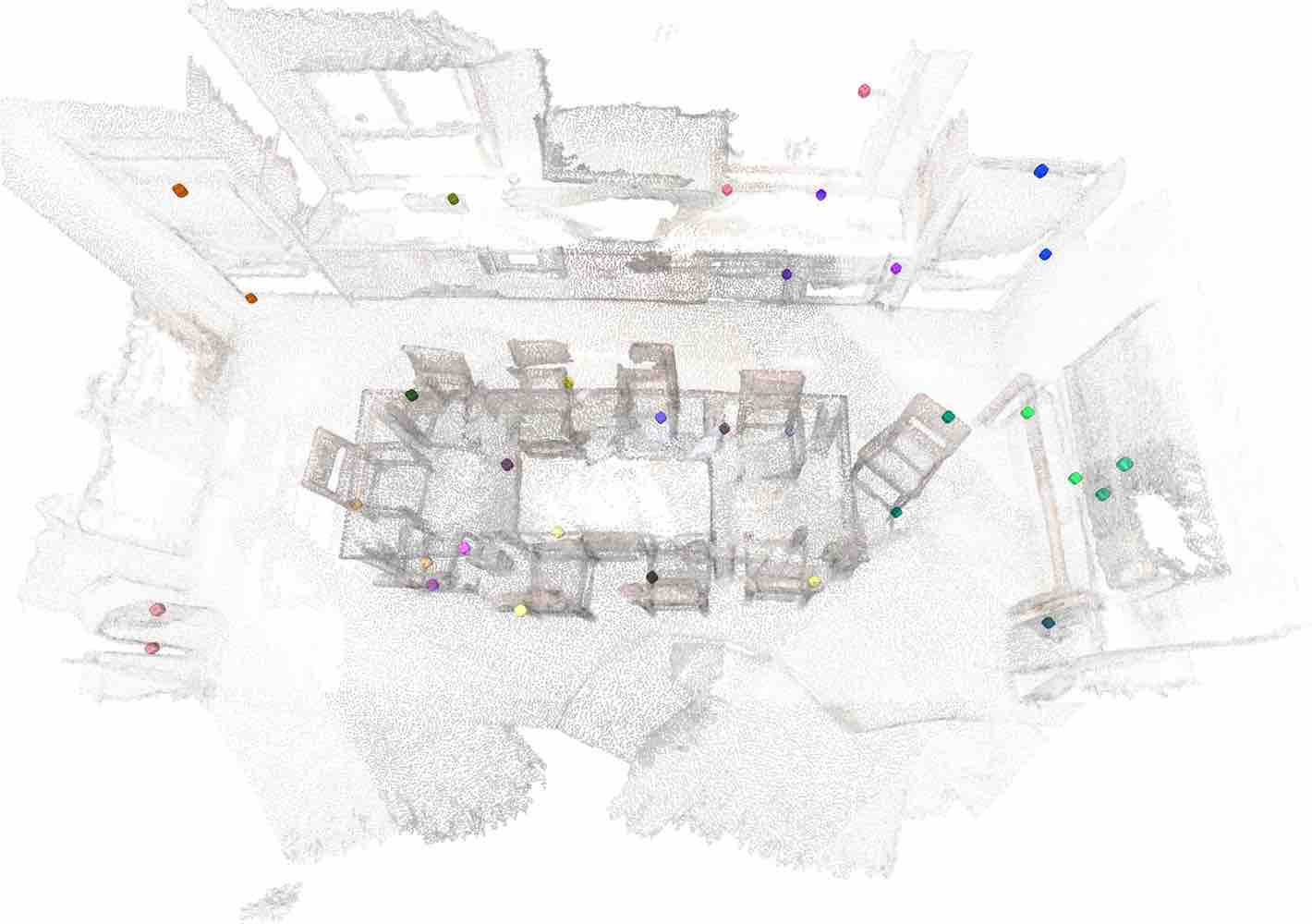} &
    \includegraphics[height=0.16\textwidth, trim={0 250px 0 0}, clip]{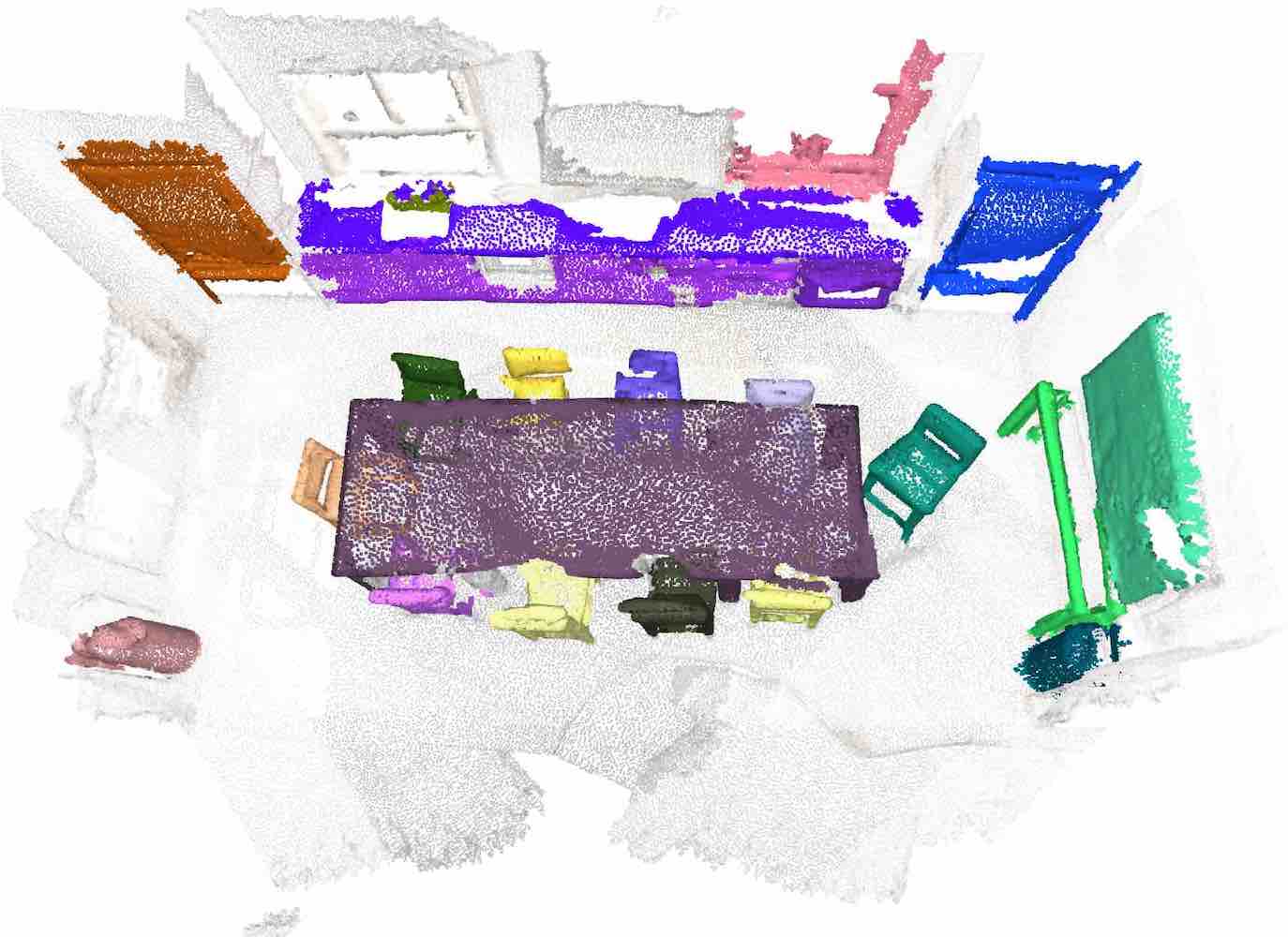} &
    \includegraphics[height=0.16\textwidth, trim={0 250px 0 0}, clip]{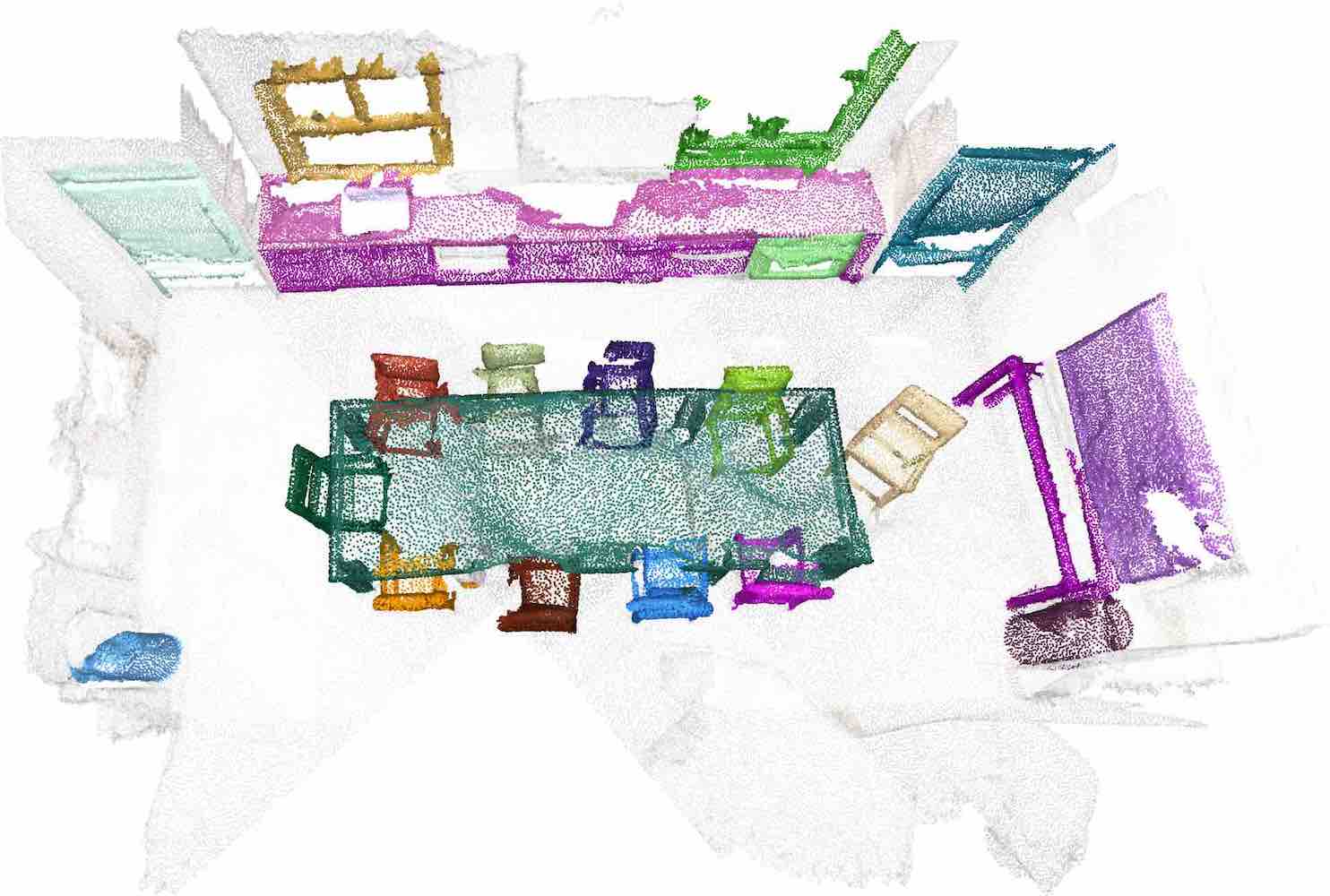} \\
Input User Clicks & Predicted Segmentation & Ground Truth Segmentation\\
\end{tabular}
\end{small}
\vspace{5px}
\caption{\textbf{Interactive 3D Object Segmentation.}
We introduce interactive object segmentation of 3D point clouds where a user collaborates hand-in-hand with a trained deep model.
\emph{Left:} A user provides sparse annotations by iteratively clicking on each object one-by-one (each color indicates a new object instance).
\emph{Middle:} Given the user clicks, the model provides instance masks that the user can then further refine.
\emph{Right:} Ground-truth for reference (Colors are picked at random so they do not necessarily match the prediction colors).}
\label{fig:teaser}
\end{figure*} 

So far, there have been few attempts to make 3D object segmentation more efficient via user interaction~\cite{Shen20eccv,Zhi21Arxiv}. In these methods, user interaction takes place in the 2D image domain, \ie{}, they additionally require images whose camera pose w.r.t.\ the point cloud is known, and associated 2D segmentation masks. In \cite{Shen20eccv} the user interacts both with 2D images and 3D point clouds (by dragging mesh vertices), forcing the user to switch between 2D and 3D views of the same object.
Due to the transition between the 2D and 3D domains,
both these methods employ custom network architectures,
therefore their integration with existing other backbones
\cite{Choy19CVPR, Thomas19ICCV, Qi17NIPS}
is not straight-forward. In contrast, our method can be integrated on more powerful future backbones or lighter ones  for AR/VR and mobile applications.
 
Here, we propose an approach where the user interacts directly with the 3D point cloud. We argue that this direct approach has several advantages: \emph{(i)} it obviates the need to supply images and their poses relative to the point cloud; \emph{(ii)} it relieves the user of going back and forth between 2D and 3D interfaces; and \emph{(iii)} it can be incorporated into any (point-based or voxel-based) 3D segmentation network.
In our experiments, we rely on MinkowskiEngine~\cite{Choy19CVPR}, proven state-of-the-art architecture based on a sparse 3D voxel grid.

A main motivation for interactive segmentation is the ability to train algorithms for new domains where no appropriate training sets are available. Our method supports this, as it is general and not tied to a specific scan pattern or class nomenclature. We train on a single dataset, ScanNet~\cite{Dai17CVPR}, and evaluate on multiple others, without retraining.

In summary, we make the following contributions:
\begin{enumerate}%
    \item First state-of-the-art for interactive object segmentation purely on 3D point data, without extra images.
    \item We study the simulation of user clicks for training purposes and introduce a practical policy to synthetically generate meaningful clicks.
    \item We experimentally show that interactive 3D object segmentation can provide accurate instance masks across various datasets, with little human effort.
    \item We show that the interactive 3D segmentation network generalizes well to different (indoor/outdoor) datasets and unseen object categories.
    \item Additionally, we have also created an interface that allows the user to pick single 3D points and used it to perform a real user study.
    \end{enumerate}

\section{Related Work}
\para{Fully-Supervised 3D Instance Segmentation.} Instance segmentation on point clouds is a fundamental task in 3D scene perception and it has been thoroughly studied in the past few years \cite{Yang19NIPS,Hou19CVPR,Engelmann20CVPR,Wang19CVPR,Jiang20CVPR,chen2021iccv,liang2021iccv}. Roughly the methods can be separated into two major categories: (1) \textit{proposal-based} and (2) \textit{clustering-based}.
Proposal-based approaches \cite{Yang19NIPS,Hou19CVPR,Engelmann20CVPR} directly generate object proposals~(inspired by Mask R-CNN~\cite{he2017mask} in 2D) and predict instance segmentation masks inside each proposal.
Clustering-based methods learn per-point features in an embedding space and then perform
clustering to generate instances\cite{Wang19CVPR,Jiang20CVPR,chen2021iccv,liang2021iccv}. Although  these  methods  have  achieved  remarkable  results  on  existing  datasets,  they  require large  amounts  of labeled  data  for  training and cannot generalize to classes that are not part of the training set. Our proposed interactive method is orthogonal to most instance segmentation methods and can be combined with them to improve segmentation masks with little user input.
\para{Weakly-Supervised learning on 3D point clouds.}
Compared to fully supervised methods, weakly supervised methods try to learn the task at hand with limited training examples or labels.
Recently, several methods~\cite{jiang2021iccv,wang2020arxiv,zhang2021iccv,liu2021cvpr,hou2021cvpr,xu2020cvpr} started to  tackle  3D  semantic  segmentation in a weakly supervised manner  aiming to achieve the performance of their fully supervised counterparts using  fewer point labels to train neural networks.
 Hou \etal{}.~\cite{houJ2021cvpr} uses contrastive learning to learn meaningful feature representations that can be fine-tuned with fewer labels.  
Liu \etal{}~\cite{liu2021cvpr} proposed a self-training approach that uses a graph propagation module that iteratively propagates the sparse labels for semantic segmentation. Xu \etal{}~\cite{xu2020cvpr} performed point cloud semantic segmentation by exploiting spatial and color smoothness constraints along with gradient approximation. 
Chibane \etal{}~\cite{chibane22eccv} learn dense instance masks from coarse bounding box labels.

While achieving impressive results,
these works focus on segmenting 3D point clouds with limited labels and sometimes require additional post-processing steps to separate object instances.
Additionally, their goal is to reach the performance of fully supervised methods using fewer labels and they do not specifically target highly accurate segmentation masks.
Even though weakly-supervised methods utilize fewer points, they still require individually trained models for every dataset.
This is unlike interactive object segmentation which provides masks for single objects enabling fast instance annotation using only a single trained model.
\para{Interactive 3D Object Segmentation.}
There is limited work on interactive frameworks for annotating 3D objects. Shen \etal{}~\cite{Shen20eccv} propose a framework for interactive 3D object annotation using interactions in both the image and the 3D domain.
First, the user provides feedback on large errors using scribbles in 2D views of the 3D space.
Then, the user corrects minor errors by manipulating mesh vertices.
The user needs to label the same object switching between two different modalities and that can be tiring for large-scale labeling.
Additionally, the method requires a specialized architecture.
\cite{valentin2015tog} also relies on handcrafted features and a custom architecture to allow users to label a scene during capture.
In contrast, our method allows users to label points directly in 3D space and can extend any semantic or instance segmentation network for 3D point clouds.
Its results shall improve when used with more powerful future architectures.
\\ \\

\vspace{-10mm}
\section{Method}
\label{sec:method}
We start by formally defining the problem. Consider a 3D scene $P \in \mathbb{R}^{N\times C}$ where $N$ is the number of 3D points in the scene and $C$ the dimensionality of the features associated with each 3D point\,(if only  $xyz$ coordinates are used: $C = 3$). 
In interactive 3D object segmentation, a model allows the user to segment the
desired region in the scene by successively placing positive
(\textit{foreground}) or negative (\textit{background}) feedback in the form of clicks on a 3D point. Every
time a new click is placed, the model outputs a new updated segmentation. Once
the user is satisfied with the  mask of the intended object, the process concludes~(Fig.~\ref{fig:inter_obj_example}).

\subsection{Input representation.}
The input to the interactive object segmentation network consists of a 3D scene $P \in \mathbb{R}^{N\times C}$ %
plus two additional channels $T_p$ and $T_n$ for the positive and negative clicks respectively.
Next, we describe these additional channels.
\para{Click Encoding Channels.} To acquire an accurate segmentation mask for the object of interest, the user provides a sequence of positive and negative clicks.
The positive clicks are considered on the object, and the negative clicks are on the background.
These clicks are represented by their 3D coordinates in the 3D scene.
$S_p$ is a set of all positive user clicks, and $S_n$ is the set of all negative ones.
Benenson \etal{}\cite{benenson19cvpr} have done an extensive study on 2D clicks encodings and find that disks of a small radius outperform the other encodings~(\eg{} distance transform) since their effects are more locally restricted~\cite{sofiiuk2021reviving}.
For those reasons,
we encode the clicks as 3D volumes.
The two additional input channels $T_p$ and $T_n$ are defined as:
\begin{equation}
T_p(p) =
\begin{cases}
     1, & \text{if } |x_p-x_q| \leq \varepsilon \text{ and}\\
     & |y_p-y_q| \leq \varepsilon \text{ and}\\
     & |z_p-z_q| \leq \varepsilon\\
     0, & \text{otherwise}
\end{cases}
\end{equation}
where $x_q \in S_p$ (set of positive clicks), and $\varepsilon$ is the volume length.
We define  $T_n$  for the negative clicks similarly with  $x_q \in S_n$ (set of negative clicks). 
So each user click channel is a binary mask of size $N \times 1$, where $N$ is the number of points,
and the mask is one for each point inside the click-volume, and zero outside.
Then we concatenate the $C$ channels of the 3D scene with $T_p$ and $T_n$ to compose a $N \times (C+2)$ input.
\para{Simulating User Clicks during Training.}
In practise, it is impractical and not realistic to collect interactive clicks from real users during training of the model.
Therefore, we simulate user clicks for training using random sampling as is common practice in the 2D domain~\cite{mahadevan18bmvc,jang19cvpr,li18cvpr,liew17iccv,xu16cvpr}.
We sample \textit{positive} clicks uniformly at random on the target 3D object.
\textit{Negative} clicks are randomly sampled from a volume around the object, up to 1.4 times bigger than the  object.
\para{Simulating User Clicks during Test.}
To ensure that the evaluation is unbiased from real user errors and users different skills,
we formalize an evaluation protocol on simulated user test clicks.
This also ensures reproducible scores.  
At test time, we add clicks one by one based on the errors of the currently predicted mask.
We imitate a user who always clicks at the center of the region with the largest error.
To obtain this error region, we need to compute pair-wise distances between all falsely labeled 3D points.
To reduce memory and computational load, it is crucial to perform this operation on a sparse voxelization of the 3D scene~(1\,point\,/\,5\,cm voxel).
\para{Online Adaptation.} Interactive object segmentation can alleviate the issue of poor generalization on unknown classes and different datasets as shown in our results on different datasets~(Tab.~\ref{tab:out_domain_evaluation}).
Importantly, the generalization can be further improved~(Tab.~\ref{tab:out_domain_evaluation}) by considering  user corrections as sparse training examples to
update the model during test~\cite{kontogianni19arxiv}.

\subsection{Network Architecture.} In this work, we adopt the Minkowski Engine~\cite{Choy19CVPR}, an implementation of sparse convolutional networks~\cite{Graham18CVPR}.
Architectures based on the Minkowski Engine have shown impressive results on 3D semantic segmentation. 
The sparse convolutions allow the efficient use of 3D convolutional layers and deep architectures already stapled in 2D vision.

All network weights except for the output layer are initialized with those provided by~\cite{Choy19CVPR}.
They are obtained by pretraining the network on the task of semantic segmentation on the ScanNetV2 dataset\,(Train)\,\cite{Dai17CVPR}.
The input now has two additional channels for the user clicks, and the last output layer is  replaced with a two-class softmax layer, which is used to produce binary segmentations~(foreground/background).
Finally, the network is fine-tuned for the task of interactive object segmentation using the ScanNetV2 train dataset adapted for foreground/background segmentation: a ground truth segmentation mask is created for each object instance where a single object instance is labeled as foreground and every other point in the 3D scene as background. 
Ground-truth segmentation masks are created for all 3D instances in the ScanNetV2 train dataset and are used to fine-tune the model on our task, that is 3D interactive object segmentation.

\subsection{Implementation and Training Details}
We set the input feature dimension $C$\,$=$\,$6$~(position and color).
For the LiDAR data of SemanticKITTI\cite{Behley19ICCV}, only 3D positions are provided, such that $C$\,$=$\,$3$.
The online adaptation experiments (Tab.~\ref{tab:out_domain_evaluation}) follow the protocol of \cite{kontogianni19arxiv} for instance adaptation~(IA) with learning rate of $5 \cdot 10^{-5} $ and $\lambda=0.1$. %

\section{Experiments}
\label{sec:experiments}

\para{Datasets.}
The proposed interactive segmentation model in all our experiments is trained on a single dataset \cite{Dai17CVPR} and tested several others \cite{Armeni16CVPR, Behley19ICCV, Dai17CVPR}.
This follows established evaluation protocol from 2D interactive object segmentation.%

In our work, we utilize the ScanNetV2~\cite{Dai17CVPR} train set since it is currently the largest of its kind containing approximately 1200 scenes. However, ScanNetV2~\cite{Dai17CVPR} is still much smaller in scale compared to the image datasets used in 2D interactive object segmentation.
To compare, in 2D interactive object segmentation, a typical deep learning model\,\cite{mahadevan18bmvc} is using the weights from pre-training on ImageNet\,\cite{deng09cvpr}, COCO\cite{lin14eccv}, and PASCAL\,VOC\, '12~\cite{pascal-voc-2012} for the task of semantic segmentation and is fine-tuned on PASCAL\,VOC\,'12~\cite{pascal-voc-2012} for the task of interactive object segmentation.
In this work, we evaluate three well-established public 3D datasets
ScanNetV2~\cite{Dai17CVPR} validation, S3DIS\,\cite{Armeni16CVPR} and SemanticKITTI\,\cite{Behley19ICCV}.%
\para{Evaluation metrics.} 
We perform the evaluation using the two standard metrics from the 2D domain~\cite{xu16cvpr,liew17iccv,mahadevan18bmvc,li18cvpr,benenson19cvpr,jang19cvpr}:
(1) \textbf{Number\,of\,Clicks\,(NOC)@q\%}, the average  number of clicks needed to reach q\% Intersection over Union\,(\iou) between predicted and \gt{} masks on every object instance~(thresholded at 20 clicks). The \underline{lower} the NOC value the \underline{better}, and (2) \textbf{\iou@k}, the average \iou{} for k number of clicks per object instance.
We additionally use the standard \textbf{AP} metric for 3D instance segmentation.

\subsection{Within-Domain Evaluation}
At first, we evaluate our approach in training and testing on the same domain. In this scenario, we assume that we would like to extend an existing dataset by labeling more scenes. For this experiment, we use the ScanNetV2 dataset~\cite{Dai17CVPR}.
We use the official split of train-validation for our experiments.
We fine-tune our model on interactive object segmentation on the training subset of ScanNetV2~\cite{Dai17CVPR} and evaluate on the validation set.
We unfortunately cannot evaluate on the test split due to hidden labels and no existing benchmark for interactive object segmentation.
ScanNetV2~\cite{Dai17CVPR} contains the segmentation masks for 40 classes. However, the ScanNetV2~\cite{Dai17CVPR} benchmark evaluates only on a 20-class subset~(18 for instance segmentation).
We use the 20-class subset of ScanNetV2~\cite{Dai17CVPR} training set to train our model, thus creating two subsets of the validation set:
(1) \textit{``seen"}, with object instances from the 20 object categories which
have been seen during training and (2) \textit{``unseen"}, with object instances of classes that have not been seen during training.
The \textit{``seen"} classes are also known as \emph{ScanNetV2 Benchmark Classes} as these are the classes evaluated on the benchmark.

\para{Results.} Results are summarized in Tables \ref{tab:comparison_w_instance} and \ref{tab:in_domain_evaluation}.
There is no previous work that uses interactive object segmentation directly on the 3D domain.
Therefore, we evaluate our method on a 3D instance segmentation setup and compare our model with state-of-the-art fully supervised methods for 3D instance segmentation (Tab.\,\ref{tab:comparison_w_instance}, \emph{top}).
For a fair comparison, we also report scores for HAIS\cite{chen2021iccv} in a class agnostic setup, where the semantic class is ignored.
When compared to fully-supervised instance segmentation methods (HAIS, SSTNet), 
our method offers segmentation masks of higher quality with only a few clicks per object.
With \textbf{5 clicks} per object, ours achieves \textbf{61.5 AP} compared to \textbf{55.1 AP} of HAIS, and with \textbf{10 clicks} the score further increase to \textbf{75.5 AP}.

\begin{table}[!t]
\vspace{5px}
\caption{
\textbf{Instance segmentation scores on ScanNetV2 validation.}}
\vspace{-5px}
\centering
\setlength{\tabcolsep}{6pt}
\begin{small}
\begin{tabular}{cp{1.8cm}cccccccc}
	\toprule
    &\textbf{Method} & & \textbf{AP} & \textbf{AP$_{50\%}$} &\textbf{AP$_{25\%}$}\\
 	\midrule
    & SSTNet\,\cite{liang2021iccv} & & $49.4$ & $64.3$ & $74.0$\\
		\parbox[t]{1mm}{\multirow{5}{*}{\rotatebox[origin=c]{90}{\textbf{Benchmark Classes}}}}&HAIS\,\cite{chen2021iccv} & & $43.5$ &$64.1$ &$75.6$\\
	\arrayrulecolor{lightgray}\cline{2-6}\arrayrulecolor{black}
	& HAIS\,\cite{chen2021iccv} & class agnostic & $55.1$ &$75.3$ &$84.9$\\
	\arrayrulecolor{lightgray}\cline{2-6}\arrayrulecolor{black}
	&\multicolumn{2}{l}{Ours\,\footnotesize{(1 click per object)}} & $20.9$ & $38.0$ & $67.2$\\
	&\multicolumn{2}{l}{Ours\,\footnotesize{(2 clicks per object)}} & $37.9$ & $60.8$ & $84.3$\\
	&\multicolumn{2}{l}{Ours\,\footnotesize{(3 clicks per object)}} & $48.1$ & $73.4$ & $90.7$\\
	&\multicolumn{2}{l}{Ours\,\footnotesize{(5 clicks per object)}} & $61.8$ & $86.0$ & $95.7$\\
	&\multicolumn{2}{l}{Ours\,\footnotesize{(10 clicks per object)}} & $75.5$ & $94.8$ & $99.0$\\
	&\multicolumn{2}{l}{Ours\,\footnotesize{(20 clicks per object)}} & $80.3$ & $96.6$ & $99.3$\\
 	\midrule
        & HAIS\,\cite{chen2021iccv} & class agnostic & $13.8$ & $26.0$ &$40.9$\\
	\arrayrulecolor{lightgray}\cline{2-6}\arrayrulecolor{black}
   		\parbox[t]{1mm}{\multirow{5}{*}{\rotatebox[origin=c]{90}{\textbf{Unseen Classes}}}}
   	&\multicolumn{2}{l}{Ours\,\footnotesize{(1 click per object)}} &  $13.5$ &  $27.8$ & $54.4$\\
	&\multicolumn{2}{l}{Ours\,\footnotesize{(2 clicks per object)}}  & $24.4$ & $45.7$ & $71.2$\\
	&\multicolumn{2}{l}{Ours\,\footnotesize{(3 clicks per object)}}  & $33.1$ & $58.4$ & $81.8$\\
	&\multicolumn{2}{l}{Ours\,\footnotesize{(5 clicks per object)}}  & $45.7$ & $75.3$ & $91.1$\\
	&\multicolumn{2}{l}{Ours\,\footnotesize{(10 clicks per object)}} & $60.3$ & $90.2$ & $97.9$\\
	&\multicolumn{2}{l}{Ours\,\footnotesize{(20 clicks per object)}} & $60.6$ & $88.3$ & $97.4$\\
    \bottomrule
\end{tabular}
\end{small}
\label{tab:comparison_w_instance}

\end{table}

The difference is even more pronounced when we evaluate on the additional \emph{``unseen"} classes of ScanNetV2~\cite{Dai17CVPR} that are not part of the training set (Tab.\,\ref{tab:comparison_w_instance}, \emph{bottom}).
ScanNetV2~\cite{Dai17CVPR} is a dataset with 40 labeled classes but usually only the 20-class subset is evaluated.
We train our model only on the 20-class benchmark subset in contrast to HAIS~\cite{chen2021iccv} which trains on all classes.
With just 2 additional clicks our model doubles the AP.
This makes our method ideal for labeling new datasets with previously unseen semantic classes. 
In summary, our method is orthogonal to instance segmentation methods. We show that it can be used to improve the quality of segmentation masks at the cost of minimal user effort.

\begin{table}[!t]
\caption{\textbf{Within-Domain Evaluation.} Interactive 3D object segmentation results on ScanNetV2 Validation}
 \vspace{-5px}
\centering
\begin{minipage}{1.0\linewidth}
	\resizebox{1.0\linewidth}{!}{%
	\setlength{\tabcolsep}{2pt}
	\begin{tabular}{p{2.2cm}rcccccccc}
	\toprule
  2D Domain  & & &  &  & \textbf{Pascal} & \textbf{Grabcut} & \textbf{Berkeley} & \multicolumn{1}{c}{\textbf{MS COCO}} & \multicolumn{1}{c}{\textbf{MS COCO}} \\
   & & &  &  &  & & & \multicolumn{1}{c}{seen classes} & \multicolumn{1}{c}{unseen classes} \\
  ($\downarrow$)\,NOC\,@\,k\% \iou{} & \multicolumn{1}{c}{} & \multicolumn{1}{c}{} & \multicolumn{1}{c}{} & \multicolumn{1}{c}{} & \multicolumn{1}{c}{$\mathbf{85\%}$} & \multicolumn{1}{c}{$\mathbf{90\%}$} &\multicolumn{1}{c}{$\mathbf{90\%}$}& \multicolumn{1}{c}{$\mathbf{85\%}$} & \multicolumn{1}{c}{$\mathbf{85\%}$} \\
\arrayrulecolor{lightgray}\midrule\arrayrulecolor{black}

	   DIOS w/ GC~\cite{xu16cvpr}& & &  &  & $6.9$ & $6.0$ & $8.7$ & \multicolumn{1}{c}{$8.3$} & \multicolumn{1}{c}{$7.8$} \\	
  	\end{tabular}}
\\
	\resizebox{\linewidth}{!}{%
	\setlength{\tabcolsep}{2pt}
	\begin{tabular}{p{2.3cm}p{1cm}p{1cm}p{1cm}|p{1cm}p{1cm}p{1cm}|p{1cm}p{1cm}p{1cm}}
		\toprule
\textbf{ScanNetV2 val-}& \multicolumn{3}{c|}{\textbf{\textit{Seen}}} & \multicolumn{3}{c|}{\textbf{\textit{Unseen}}}& \multicolumn{3}{c}{\textbf{All}}\\
         NOC @ k \%~\iou{}& 
         \multicolumn{1}{c}{$\mathbf{80\%}$} &
         \multicolumn{1}{c}{$\mathbf{85\%}$} & 
         \multicolumn{1}{c|}{$\mathbf{90\%}$}&
         \multicolumn{1}{c}{$\mathbf{80\%}$} & 
         \multicolumn{1}{c}{$\mathbf{85\%}$} & 
         \multicolumn{1}{c|}{$\mathbf{90\%}$}&
         \multicolumn{1}{c}{$\mathbf{80\%}$}& 
         \multicolumn{1}{c}{$\mathbf{85\%}$} & 
         \multicolumn{1}{c}{$\mathbf{90\%}$} \\
        \arrayrulecolor{lightgray}\midrule\arrayrulecolor{black}

	Ours &  
 \multicolumn{1}{c}{$8.3$}  &   
 \multicolumn{1}{c}{$10.6$}  &
 \multicolumn{1}{c|}{$13.6$} &
 \multicolumn{1}{c}{$11.7$}  &
 \multicolumn{1}{c}{$14.1$}&
 \multicolumn{1}{c|}{$16.5$} &
 \multicolumn{1}{c}{$11.6$}  &
 \multicolumn{1}{c}{$14.1$} &
 \multicolumn{1}{c}{$14.7$} \\
        \bottomrule
	\end{tabular}
	}
	\end{minipage}
 \label{tab:in_domain_evaluation}
\end{table}

Our evaluation on  interactive 3D object segmentation is presented in Tab.~\ref{tab:in_domain_evaluation}.
To provide a frame of reference (since there are no previous methods on interactive 3D object segmentation) we report the scores~(Tab.~\ref{tab:in_domain_evaluation}) from the first deep learning paper~\cite{xu16cvpr} on 2D interactive object segmentation.
Our scores are comparable with its 2D counterpart even though the 2D method uses networks pre-trained on much bigger datasets and some of the evaluation datasets contain only one object instance per image~(\eg{}, GrabCut~(\cite{Rother04-tdfixed}) compared to the 30 objects of an average indoor 3D scene.
The 2D method also employs additional GraphCut\cite{xu16cvpr} refining. %

We additionally report our scores compared to the method of Shen \etal{}~\cite{Shen20eccv} that segments objects using 2D scribbles in Tab.\ref{tab:scribbles}.
\cite{Shen20eccv} reports scores on selected classes of the Pix3D~\cite{pix3d} dataset.
Annotators labelled 95 randomly-selected objects to be used for fine-tuning a 3D reconstruction model pre-trained on synthetic data.
Additionally, it is not easy to define the size and form of the scribbles used.
The labeled objects were un-occluded and un-truncated so their statistics are not directly comparable with our evaluation on the densely populated datasets of ScanNetV2~(val) and S3DIS.
However, the comparison might show a trend for common object classes.

\begin{table}
\vspace{5px}
\caption{Comparison with Contrastive Scene Context (CSC)\cite{hou2021cvpr}.}
\vspace{-5px}
    \centering
    \begin{tabular}{lcccc}
    \toprule
     & \textbf{Points} & \textbf{Points} & \textbf{Objects} & \\%\relax
    \textbf{Method} & \textbf{per Scene} & \textbf{per Objects} & \textbf{per Scene} & \textbf{AP$_{50\%}$} \\%\relax
     \midrule
      CSC\cite{hou2021cvpr}  &  $200$ (train) & -& - & $48.9$\\
      Ours & $60$ (test) & $2$  & $\approx 30$   & $\textbf{60.8}$\\
      Ours & $90$ (test) & $3$  &  $\approx 30$  & $\textbf{73.4}$\\
    \bottomrule
    \end{tabular}
    \label{tab:semi}
\end{table}
\para{Comparison with Weakly-Supervised.} We compare our method with the weakly supervised method of~\cite{hou2021cvpr} in Tab.~\ref{tab:semi}. Weakly-supervised methods aim to achieve fully supervised accuracy with less labeled data, while our method on the other hand aims to supersede the performance of existing fully supervised methods~(as shown in Fig.~\ref{fig:iou_level}, labeled datasets require segmentation mask of the highest quality) using limited additional human input. Unlike weakly-supervised methods, our approach does not require re-training the model (even with reduced number of points)
when we want the segmentation masks of classes not existing in the training set or when used in different datasets. Our method performs significantly better than~\cite{hou2021cvpr} with less input from the user.
Importantly, unlike weakly-supervised methods, our method does not require re-training when new user inputs appear.

\begin{table}[!b]
\caption{
\textbf{Out-of-Domain Evaluation.}
Interactive 3D object segmentation scores on S3DIS and SemanticKITTI.}
\vspace{-5px}
\centering
\setlength{\tabcolsep}{6pt}
\begin{tabular}{p{2.3cm}ccc|ccc}
\toprule
\emph{(trained on ScanNet)}	 & \multicolumn{3}{c|}{\textbf{S3DIS Area 5}\cite{Armeni16CVPR}}
	 & \multicolumn{3}{c}{\textbf{SemanticKITTI}\cite{Behley19ICCV}}\\
NOC @ k \% \iou{}
    & \multicolumn{1}{c}{$\mathbf{80\%}$}
    & \multicolumn{1}{c}{$\mathbf{85\%}$}
    & \multicolumn{1}{c|}{$\mathbf{90\%}$}
    &\multicolumn{1}{c}{$\mathbf{80\%}$}
    & \multicolumn{1}{c}{$\mathbf{85\%}$}
    & \multicolumn{1}{c}{$\mathbf{90\%}$} \\
\arrayrulecolor{lightgray}\midrule\arrayrulecolor{black}
Ours  &  $6.8$  & $8.9$ &  $11.8$ & $19.4$ & $19.5$ & $19.6$\\
		Ours (online adapt.) &  6.5  & 8.6   & 11.5   & 15.1 & 15.4 & 15.9\\
        \bottomrule
	\end{tabular}
 	\label{tab:out_domain_evaluation}

\end{table}

\subsection{Out-of-Domain Evaluation.}
In this section, we evaluate the performance of our method on datasets different from our training dataset.
We would like to see the performance of our model both in relatively small distribution shifts between training and test using two indoor datasets.
(ScanNet $\xrightarrow[]{}$ S3DIS).
Additionally, we evaluate on large distribution shifts (ScanNet $\xrightarrow[]{}$ SemanticKITTI) by using our model trained on a indoor dataset and testing on an outdoor dataset containing different objects and recorded with a different depth sensor like SemanticKITTI~\cite{Behley19ICCV}).
First, we use our model trained on ScanNetV2 training set \emph{without any fine-tuning} on these datasets.
We additionally report separate scores when online adaptation is used.

\para{Results.}
Results are summarized in Tables~\ref{tab:out_domain_evaluation} and \ref{tab:S3DISperclass}.
As shown in Tab.~\ref{tab:out_domain_evaluation} (row\,1) even though our interactive segmentation network was trained only on the ScanNet dataset it can still generalize easily to other indoor datasets like S3DIS.
It even requires less clicks to achieve segmentation masks of the same quality.
The more our test dataset distribution differs from the training, the more clicks are required to achieve good segmentation mask as shown in the results on semanticKITTI~(Tab.~\ref{tab:out_domain_evaluation}).
The SemanticKITTI dataset requires more clicks for two reasons:
(1) its sparsity restricts the needed context from the click masks and
(2) its characteristics (sparsity, no color, different sensor capturing the data) are very different from the training dataset.
To that end, we incorporate an \emph{online adaptation} method that updates the model weights during test time using sparse user annotations~\cite{kontogianni19arxiv}.
Tab.~\ref{tab:out_domain_evaluation} (row\,2) shows that online adaptation helps significantly to adapt to new dataset distributions, requiring 4 less clicks.
The adaptation does not help significantly on the S3DIS dataset since ScanNet and S3DIS do not differ as much as ScanNet and SemanticKITTI.
However, if we look at the evaluation metrics per class~(the method is class agnostic, we only aggregated the results per individual class), online adaptation helps significantly to segment objects belonging to the classes of \textit{board} and \textit{beam}~(Tab.~\ref{tab:out_domain_evaluation}), \ie{} two classes that do not exist in this form in ScanNet.

\begin{table}[!t]
\vspace{5px}

\caption{{Per-class interactive 3D object segmentation}~(S3DIS A5~\cite{Armeni16CVPR})}
\vspace{-5px}
\setlength\tabcolsep{3pt}
\resizebox{0.99\linewidth}{!}{
\begin{tabular}{lc gggccccccccss}
\toprule

NOC@90 \\Method	  & \rotatebox{90}{mean} &
\rotatebox{90}{Ceiling}&%
\rotatebox{90}{Floor}&%
\rotatebox{90}{Wall}&%
\rotatebox{90}{Clutter}&%
\rotatebox{90}{Column}&%
\rotatebox{90}{Window}&%
\rotatebox{90}{Door}&%
\rotatebox{90}{Table}&%
\rotatebox{90}{Chair}&%
\rotatebox{90}{Bookcase}&%
\rotatebox{90}{Sofa}&%
\rotatebox{90}{Board}&%
\rotatebox{90}{Beam}\\
\midrule
Ours	& $11.8$	& $4.6$	& $2.1$	& $11.4$	&$15.0$ 	& $11.9$	& $11.5$	& $13.3$	& $11.8$	& $5.3$	& $10.6$	& $5.7$	& $16.0$	& $17.7$	\\
Ours w/ \cite{kontogianni19arxiv} 		& $11.5$	& $4.5$	& $2.2$	& $11.0$	& $14.8$	& $11.4$	& $10.5$	& $12.9$	& $11.3$	& $5.2$	& $10.2$	& $5.0$	& $14.7$	& $14.0$	\\

\bottomrule
\end{tabular}
}
\label{tab:S3DISperclass}

\end{table}

\begin{table}[!b]
	\caption{\textbf{3D clicks compared to 2D scribbles.}}
\vspace{-5px}
	\centering
    \begin{minipage}{1.0\linewidth}
	\centering
	\label{tab:3dvs2dscribbles}
	\resizebox{1.0\linewidth}{!}{%
	\setlength{\tabcolsep}{3pt}

	\begin{tabular}{p{0.9cm}ccccccccc}
	\toprule
    & \textbf{Annotations} & \textbf{Dataset} &\textbf{Bed} & \textbf{Bookc.} &\textbf{Desk} &  \textbf{Sofa} & \textbf{Table}& \textbf{Wardr.} \\
    \arrayrulecolor{lightgray}\hline\arrayrulecolor{black}

   SIM~\cite{Shen20eccv} & 2D scribbles &Pix3D& 15 & 18 &10 & 10 & 19 & 9 \\ 
   Ours   & 3D clicks  & S3DIS & -  &10.6  & -  & 5.7 & 11.8 & - \\ 
   Ours & 3D clicks & ScanNet\,val. & 16.3 & 15.7 &18.5 & 12.5 & 11.8 & - \\ 
    \bottomrule
	\end{tabular}
    }
	\end{minipage}
\label{tab:scribbles}	
 \end{table}

\subsection{Qualitative Results.} 
\begin{figure*}[!htbp]
\vspace{5pt}
\centering
\begin{tabular}{p{0.1cm}cccc}
	\setlength{\tabcolsep}{3pt}

\rotatebox{90}{\small \hspace{3mm}ScanNet\,\cite{Dai17CVPR}}&
    \begin{overpic}[trim={0cm 0cm 0cm 1cm},clip, width=0.18\linewidth]{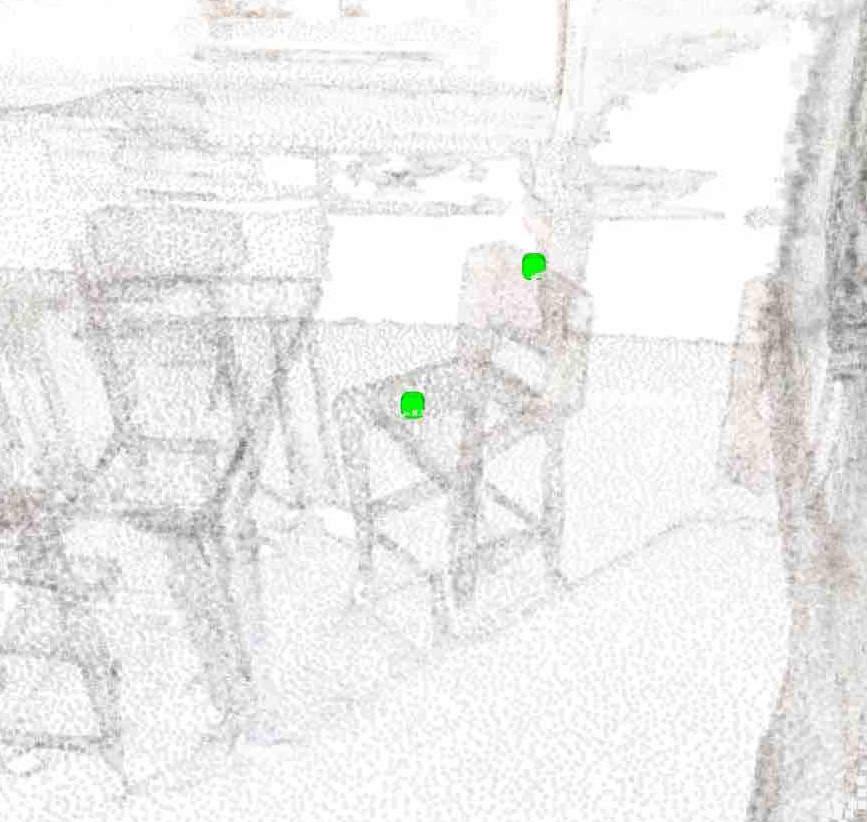}
\put(200,130){\makebox(0,-10){\color{black}\textbf{\scriptsize k=2}}\color{black}}
\end{overpic} &    
\begin{overpic}[trim={0cm 0cm 0cm 1cm},clip, width=0.18\linewidth]{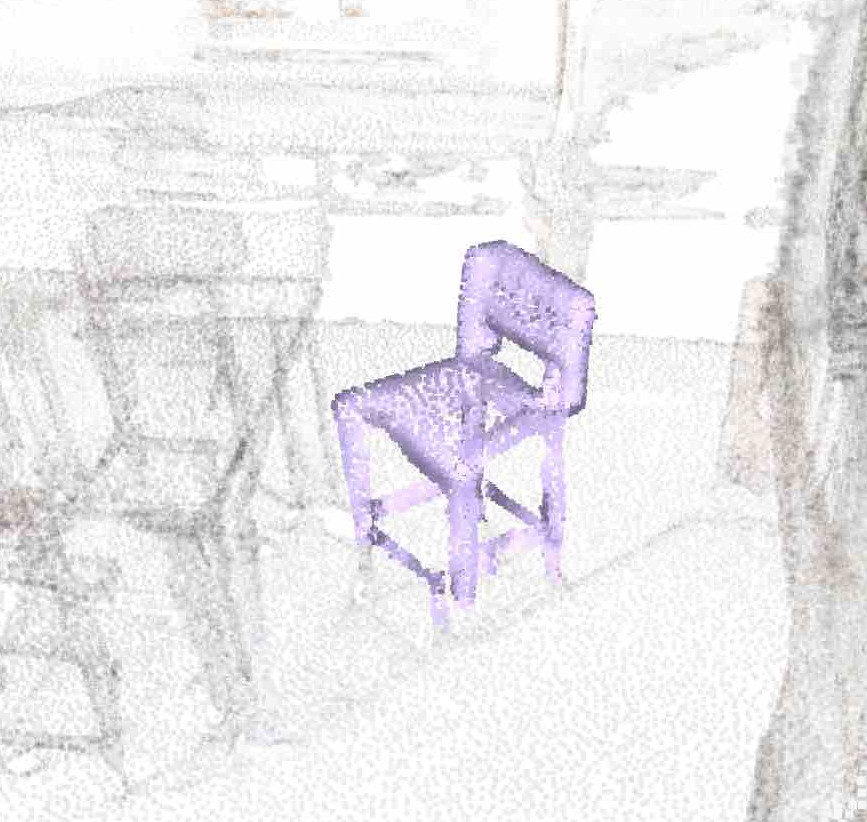}
\put(300,100){\makebox(0,-10){\color{black}\textbf{\scriptsize IOU=95.8}}\color{black}}
\end{overpic} &
\begin{overpic}[trim={0cm 0cm 0cm 2.2cm},clip, width=0.18\linewidth]{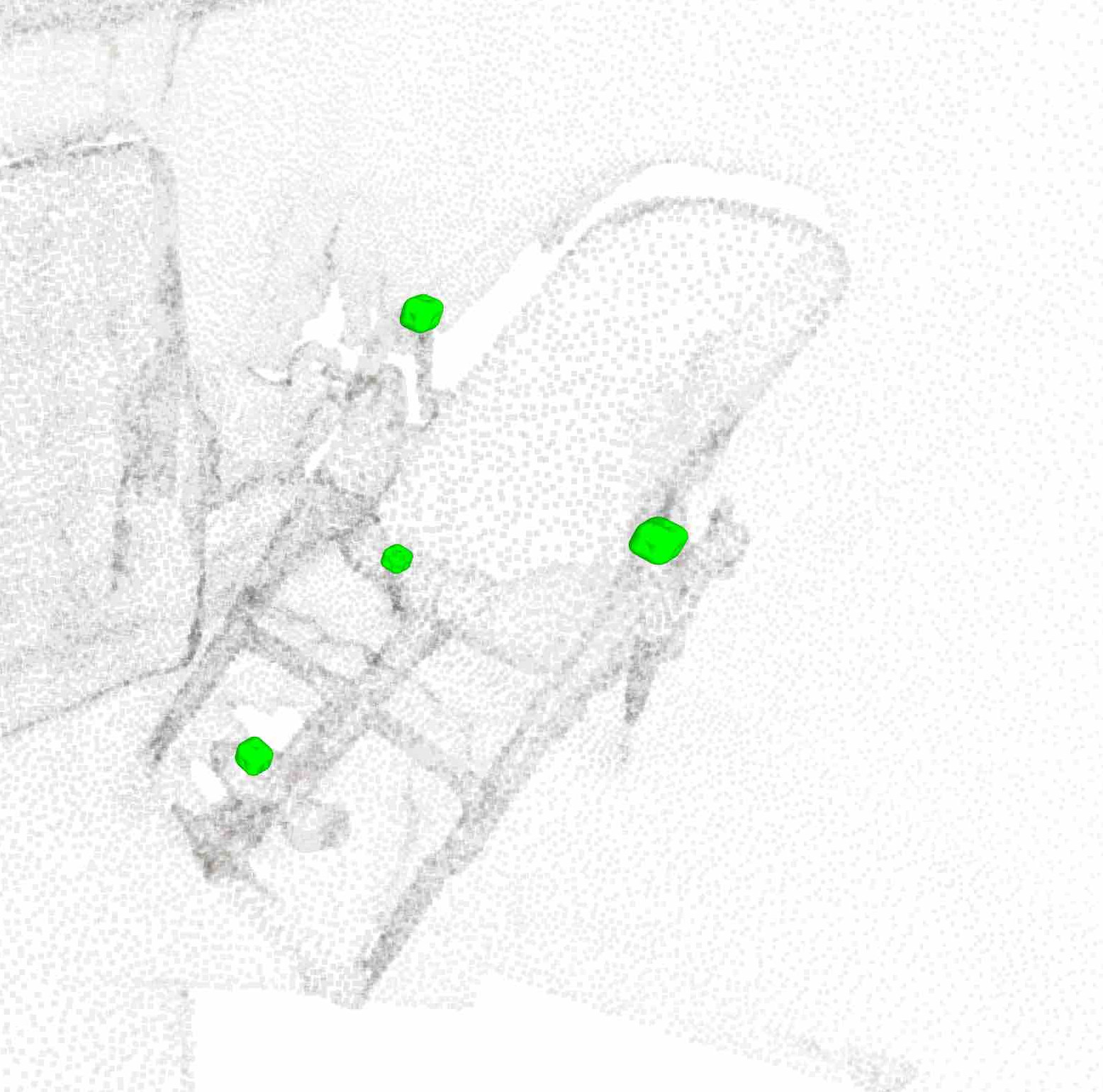}\put(200,140){\makebox(0,-10){\color{black}\textbf{\scriptsize k=4}}\color{black}} 
\end{overpic} &
\begin{overpic}[trim={0cm 0cm 0cm 2.2cm},clip, width=0.18\linewidth]{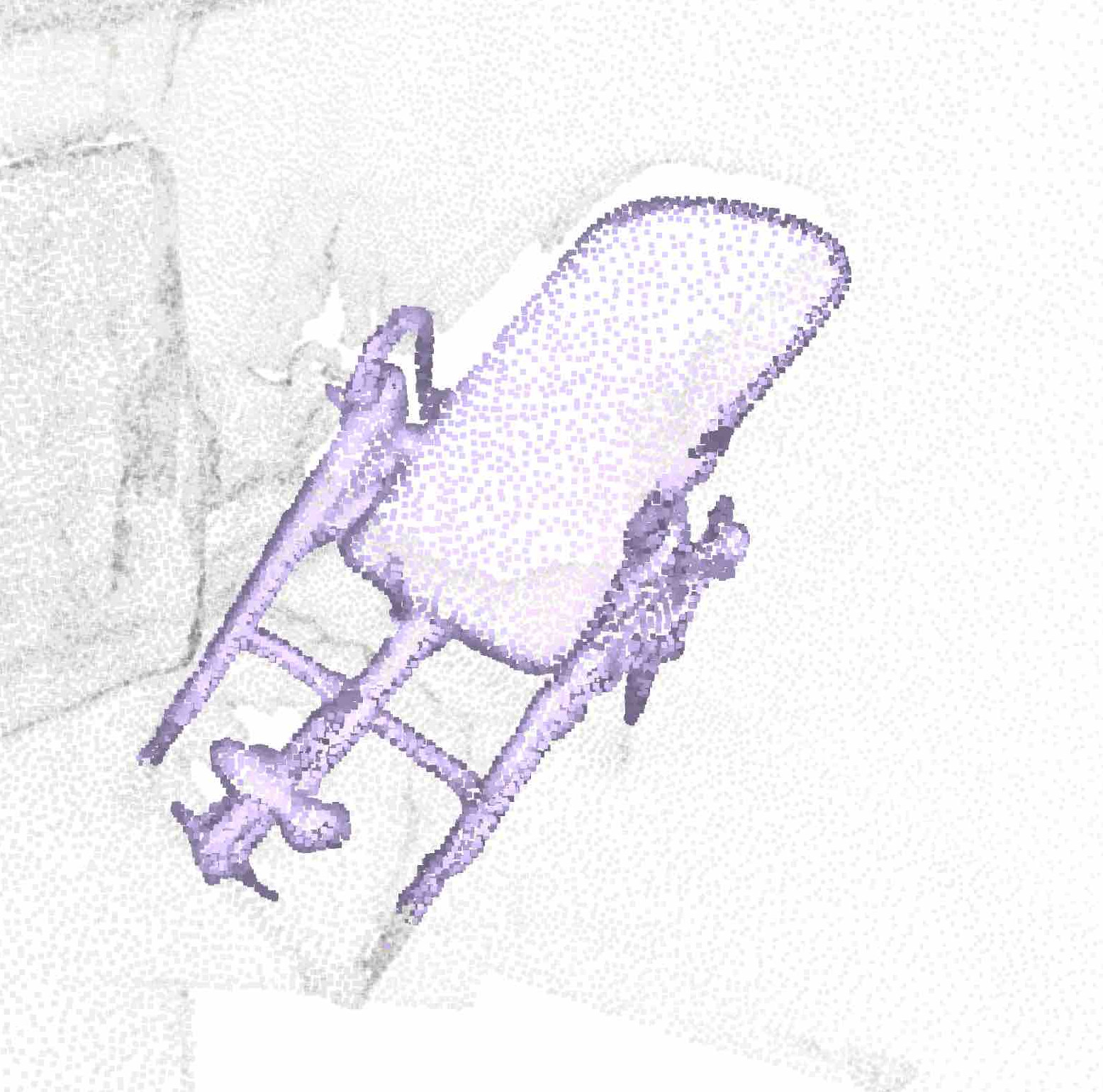}\put(300,100){\makebox(0,-10){\color{black}\textbf{\scriptsize IOU=95.6}}\color{black}}
\end{overpic}
\\
\rotatebox{90}{\small \hspace{4mm}ScanNet*\,\cite{Dai17CVPR}}&
\begin{overpic}[trim={4cm 2cm 0cm 4cm},clip, height=0.15\linewidth]{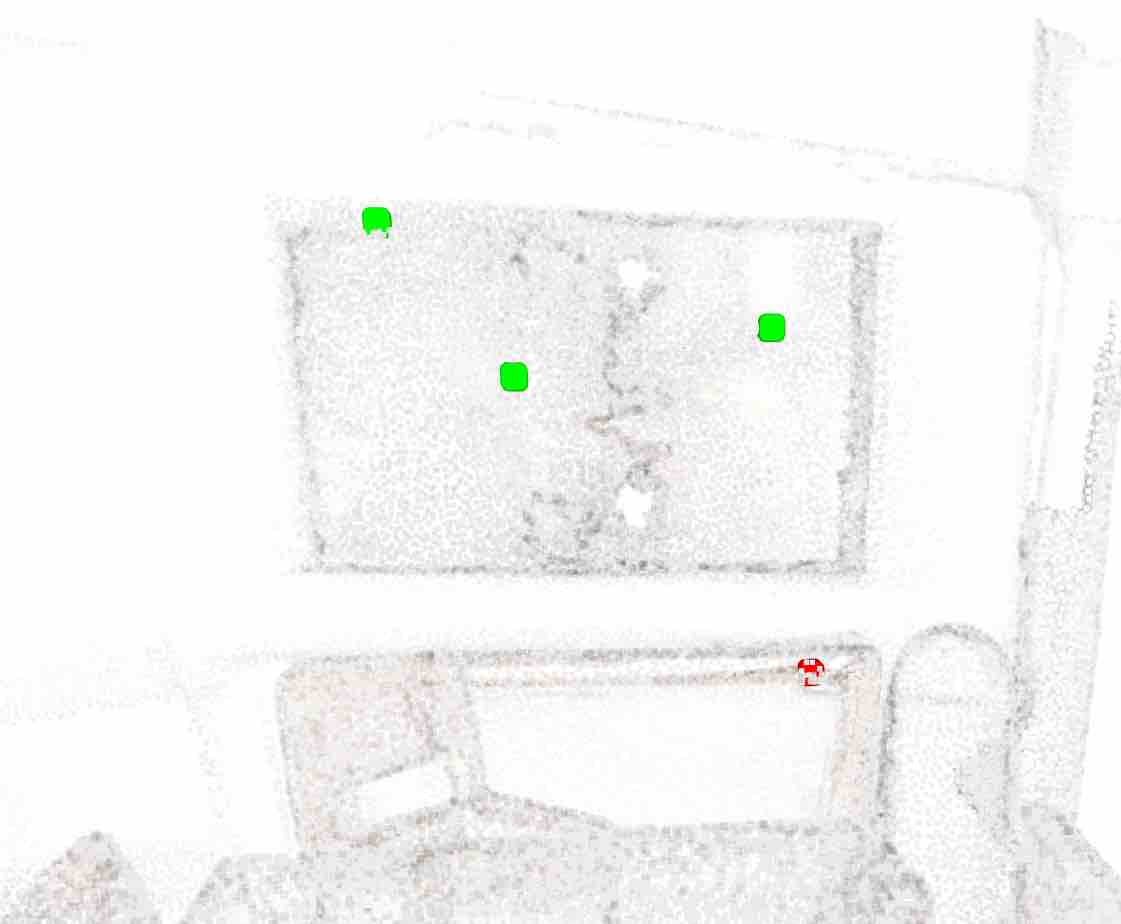}
\put(200,140){\makebox(0,-10){\color{black}\textbf{\scriptsize k=4}}\color{black}}
\end{overpic} &    
\begin{overpic}[trim={4cm 2cm 0cm 0cm},clip, height=0.15\linewidth]{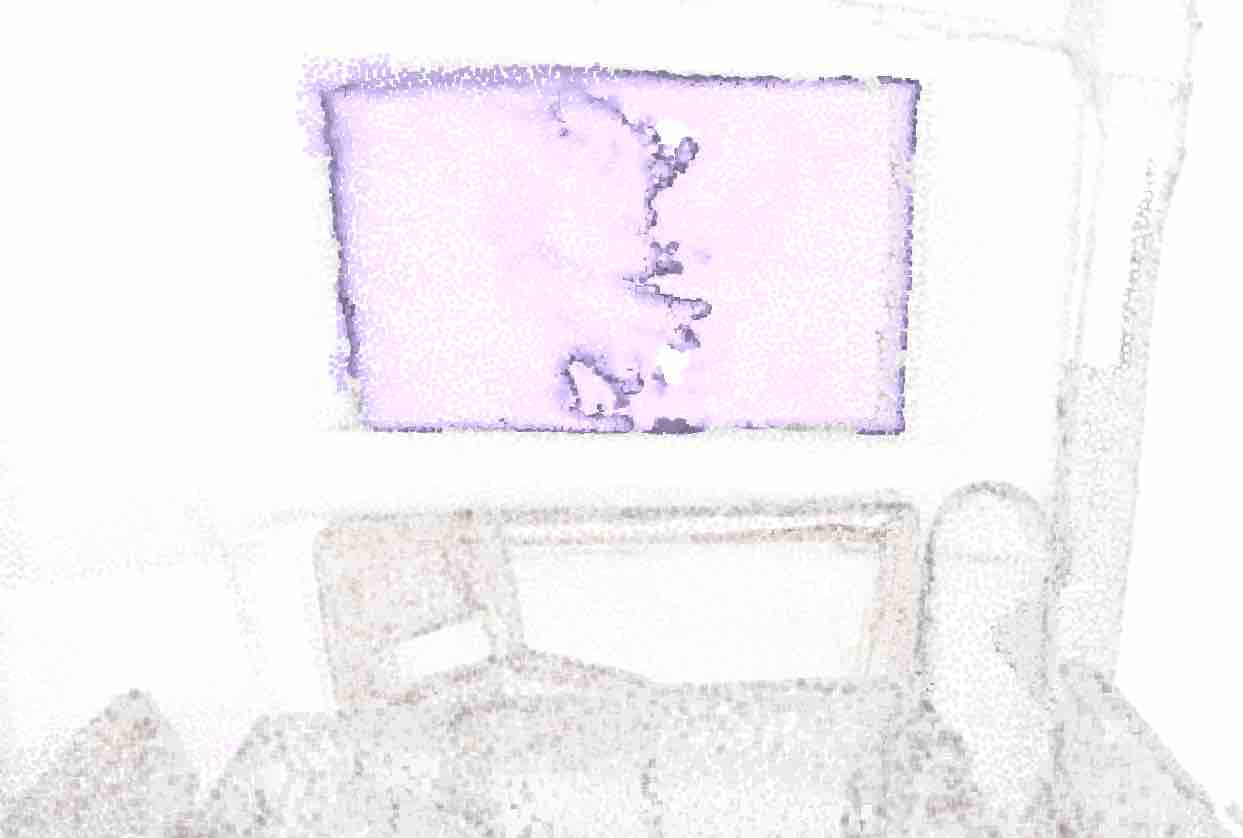}
\put(300,100){\makebox(0,-10){\color{black}\textbf{\scriptsize IOU=89.2}}\color{black}}
\end{overpic} &
\begin{overpic}[trim={2cm 2cm 0cm 6cm},clip, height=0.15\linewidth]{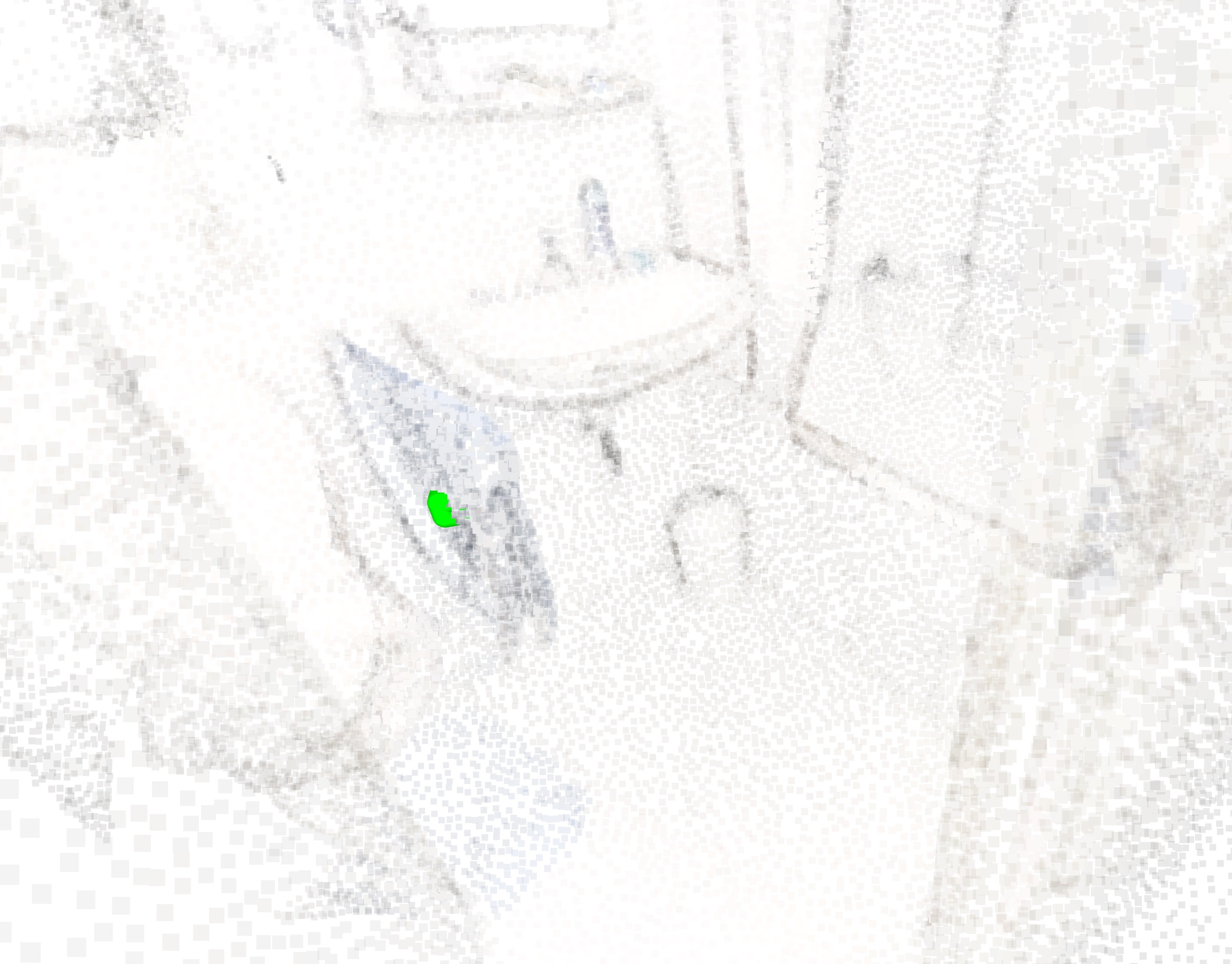}\put(200,140){\makebox(0,-10){\color{black}\textbf{\scriptsize k=1}}\color{black}}
\end{overpic} &
\begin{overpic}[trim={2cm 2cm 0cm 6cm},clip, height=0.15\linewidth]{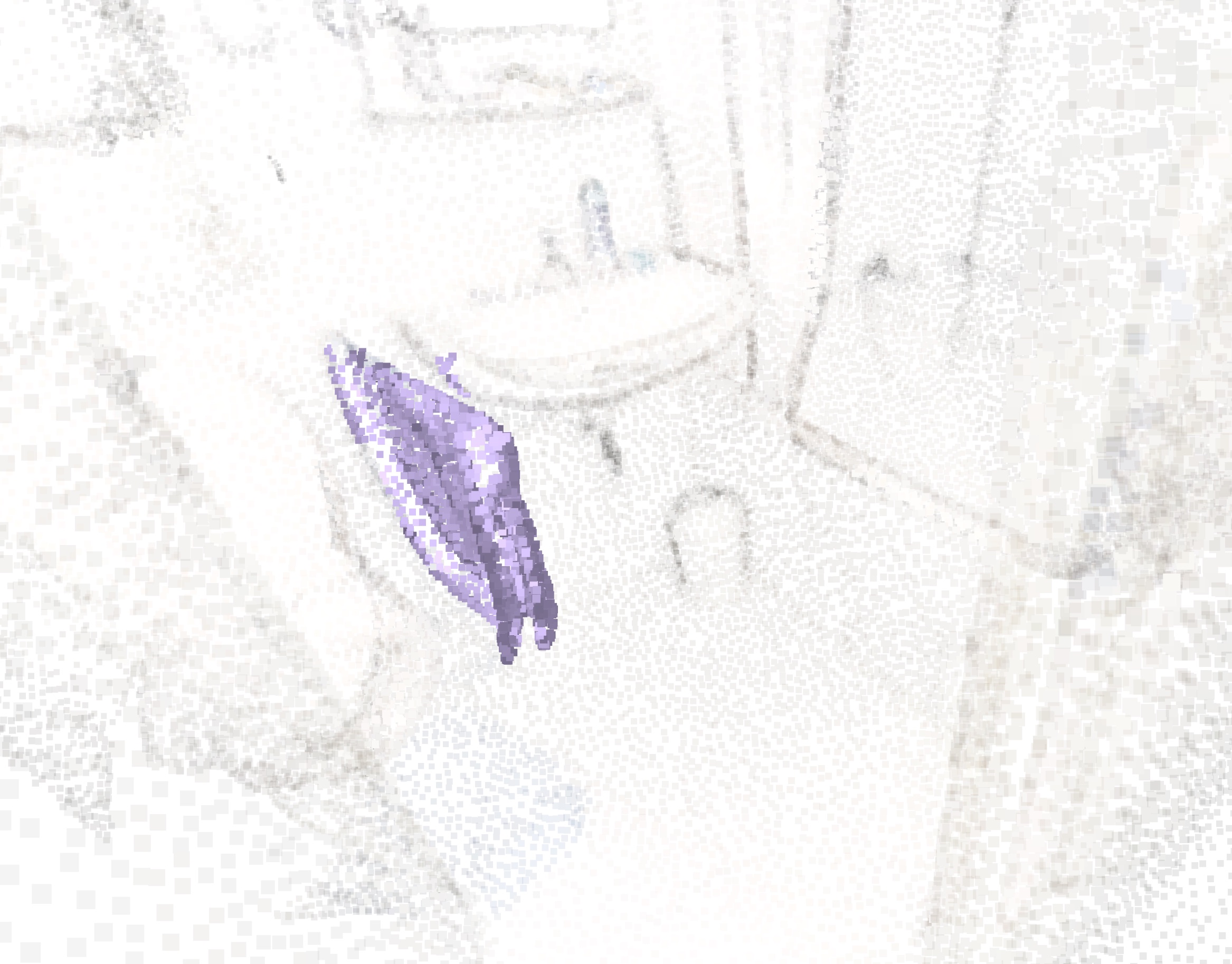}\put(300,100){\makebox(0,-10){\color{black}\textbf{\scriptsize IOU=80.7}}\color{black}}
\end{overpic}
\\
\rotatebox{90}{\small \hspace{0.7cm}S3DIS\,\cite{Armeni16CVPR}}&
\begin{overpic}[trim={0cm 0cm 0cm 0cm},clip, height=0.15\linewidth]{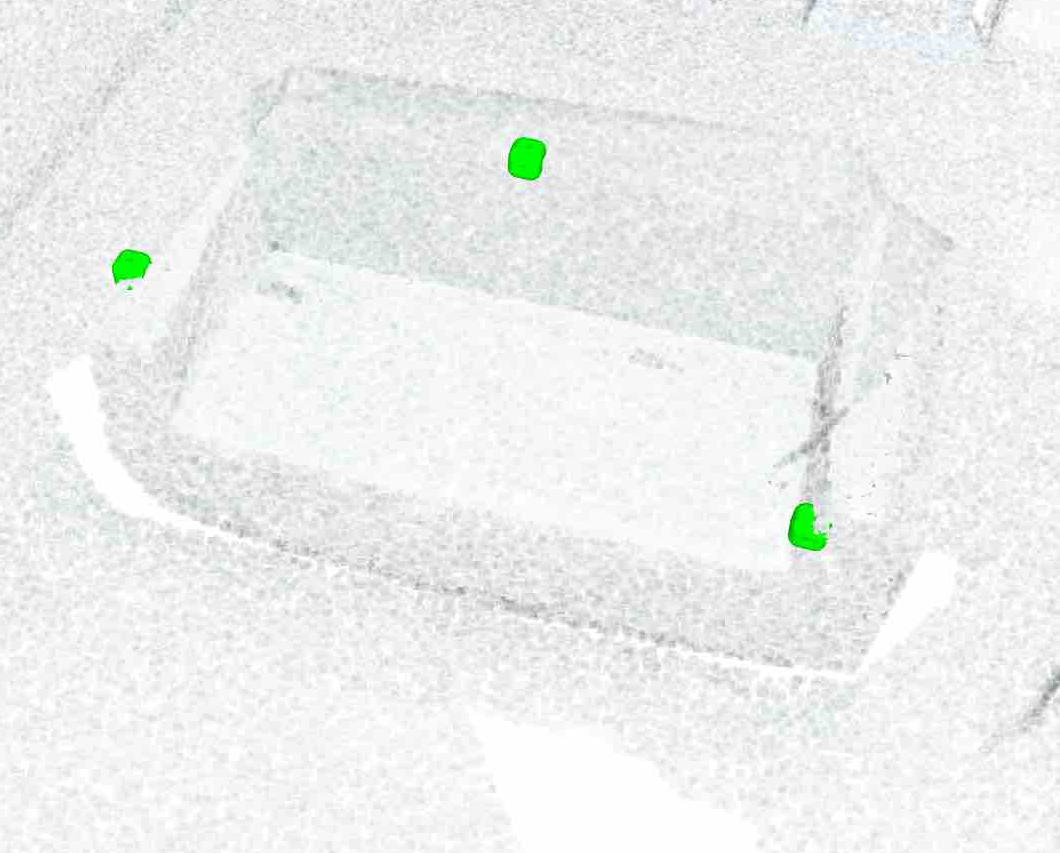}
\put(200,100){\makebox(0,-10){\color{black}\textbf{\scriptsize k=3}}\color{black}}
\end{overpic} &
\begin{overpic}[trim={0cm 0cm 0cm 0cm},clip, height=0.15\linewidth]{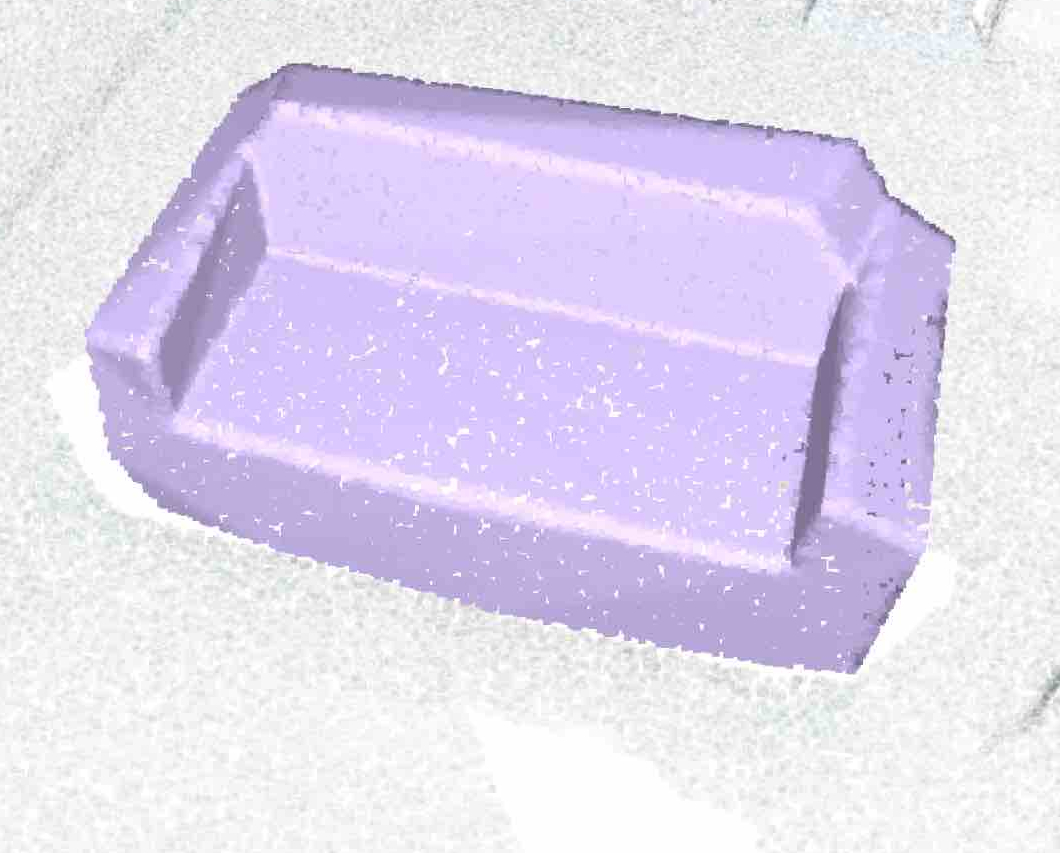}
\put(300,100){\makebox(0,-10){\color{black}\textbf{\scriptsize IOU=99.2}}\color{black}}
\end{overpic}&
\begin{overpic}[trim={1cm 0cm 1cm 0cm},clip, height=0.15\linewidth]{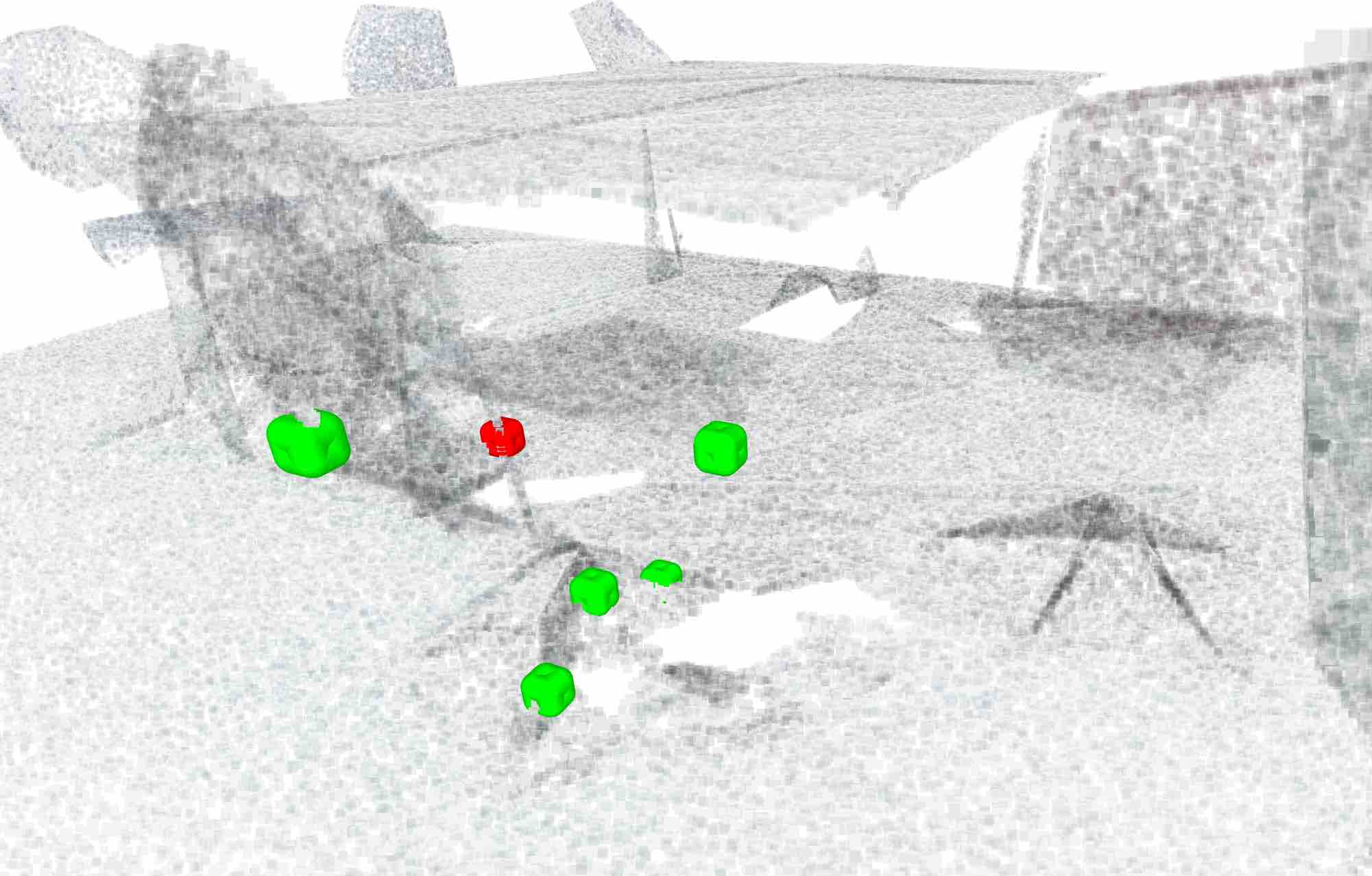}\put(200,90){\makebox(0,-10){\color{black}\textbf{\scriptsize k=6}}\color{black}}
\end{overpic} &
\begin{overpic}[trim={1cm 0cm 1cm 0cm},clip, height=0.15\linewidth]{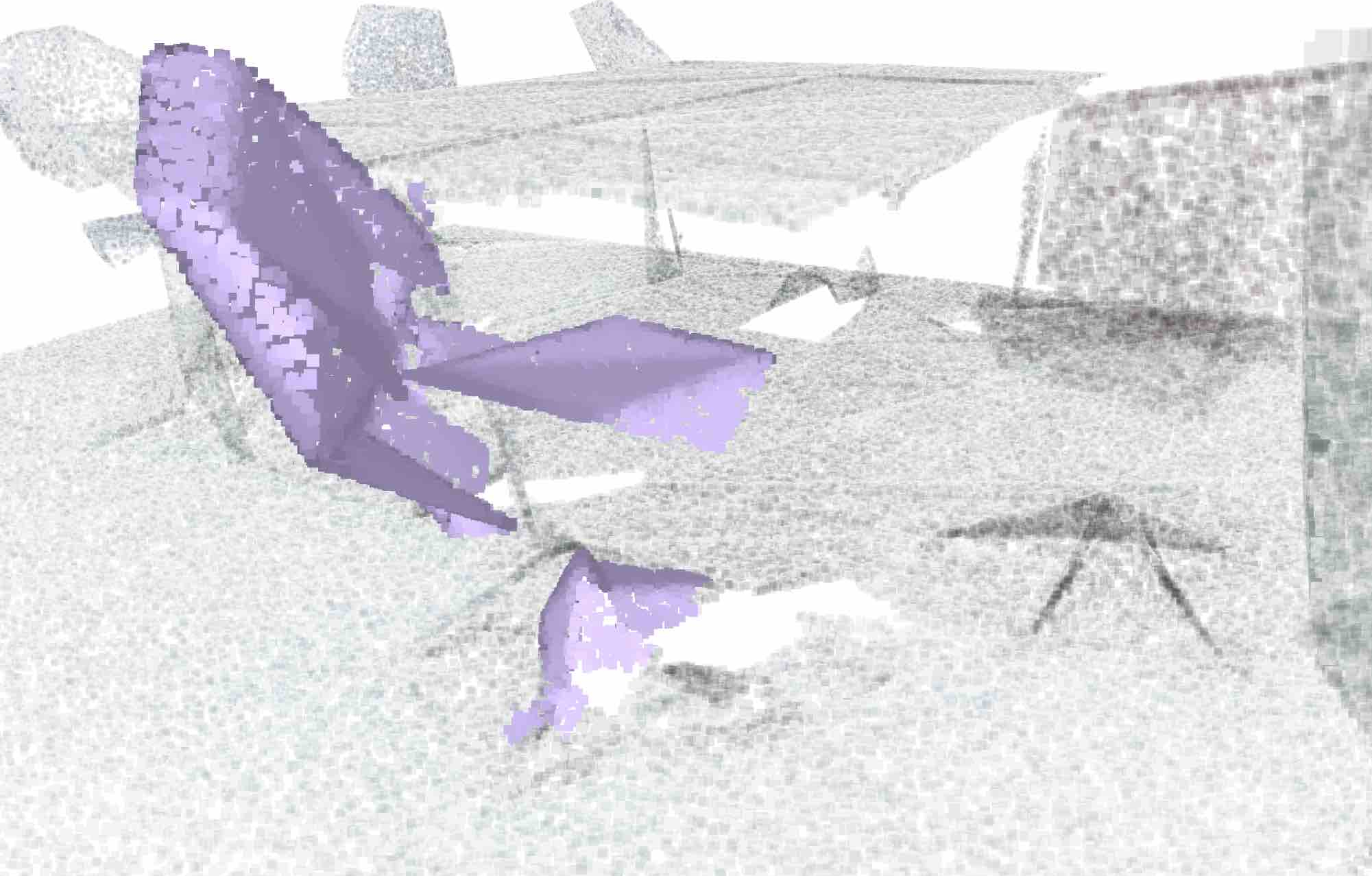}\put(300,130){\makebox(0,-10){\color{black}\textbf{\scriptsize IOU=90.7}}\color{black}}
\end{overpic}\\

\rotatebox{90}{\small SemanticKITTI\,\cite{Behley19ICCV}}&
\begin{overpic}[trim={0cm 0cm 0cm 0cm},clip, width=0.2\linewidth]{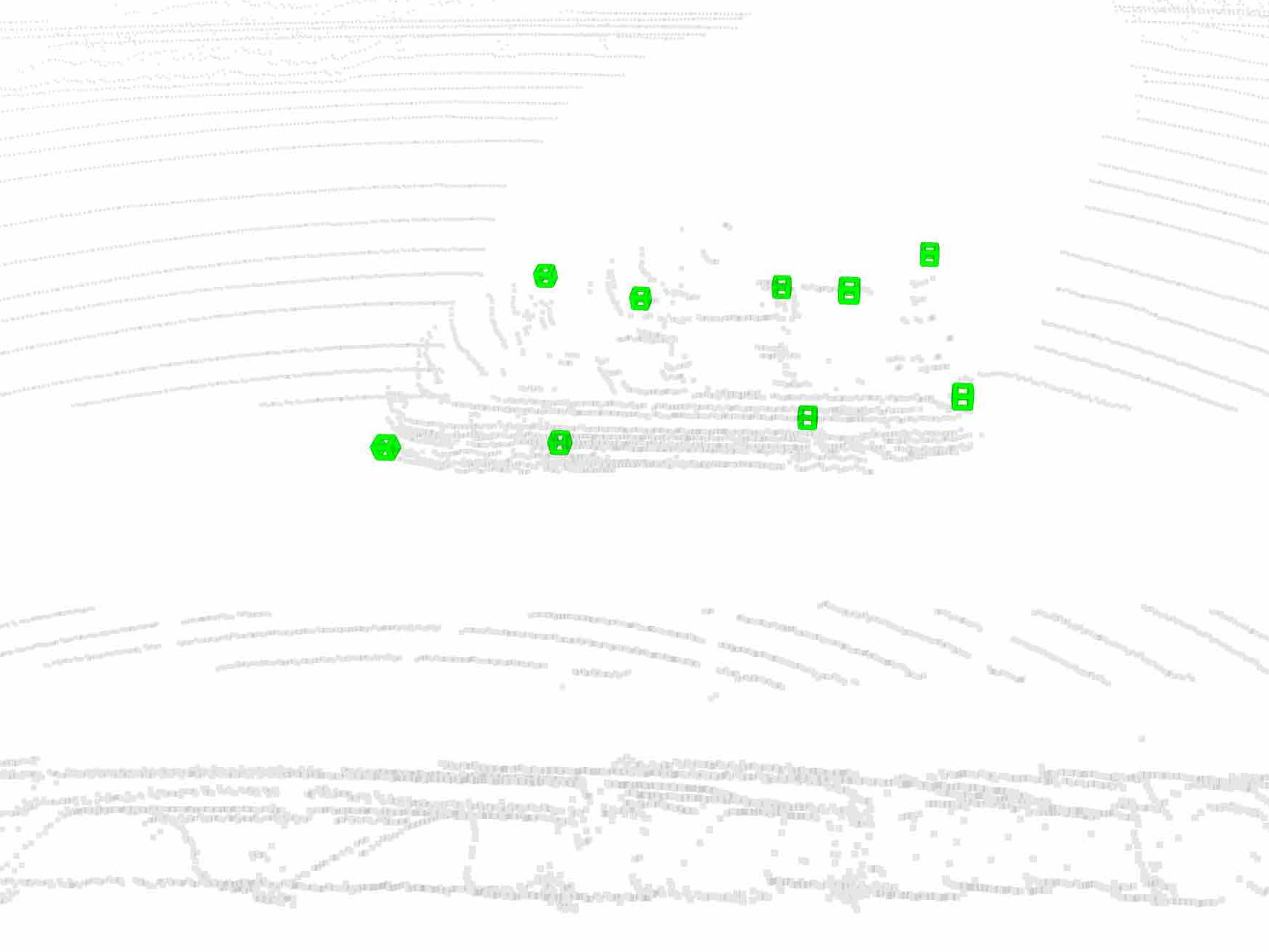}
\put(200,100){\makebox(0,-10){\color{black}\textbf{\scriptsize k=9}}\color{black}}
\end{overpic} &
\begin{overpic}[trim={0cm 0cm 0cm 0cm},clip, width=0.2\linewidth]{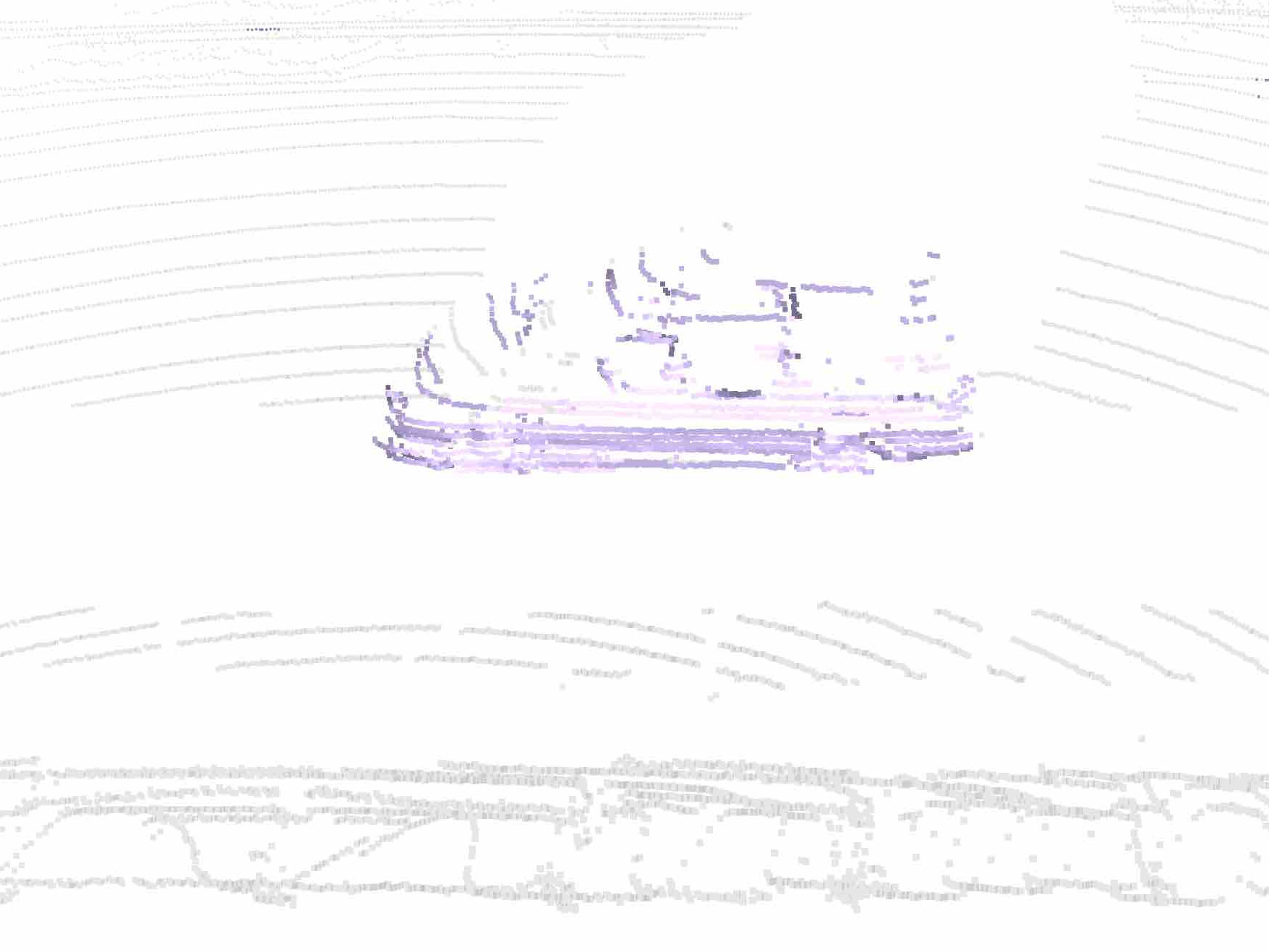}
\put(300,100){\makebox(0,-10){\color{black}\textbf{\scriptsize IOU=89.4}}\color{black}}
\end{overpic} &
\begin{overpic}[trim={0cm 0cm 0cm 0cm},clip, width=0.2\linewidth]{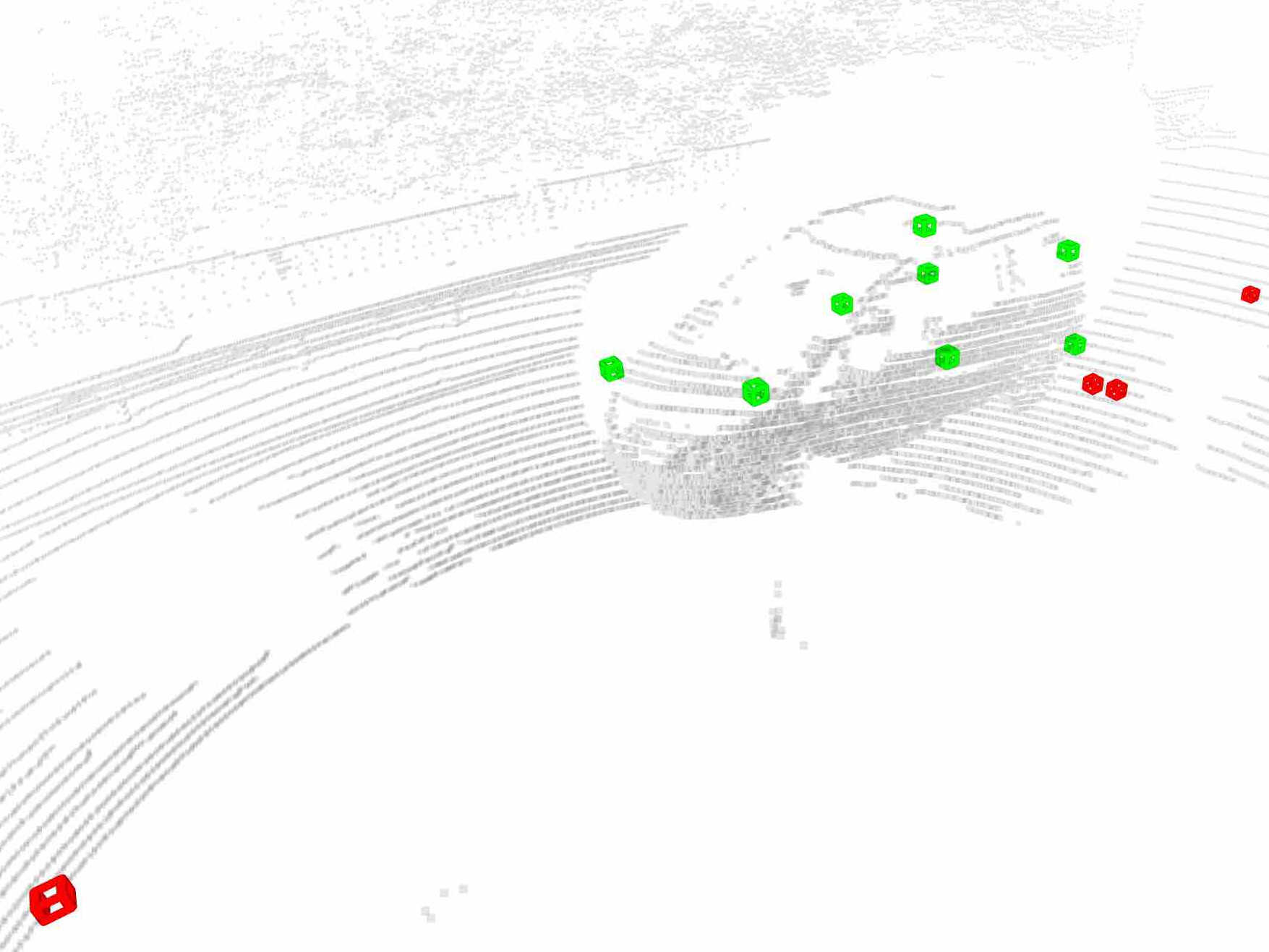}\put(200,100){\makebox(0,-10){\color{black}\textbf{\scriptsize k=16}}\color{black}}
\end{overpic} &
\begin{overpic}[trim={0cm 0cm 0cm 0cm},clip, width=0.2\linewidth]{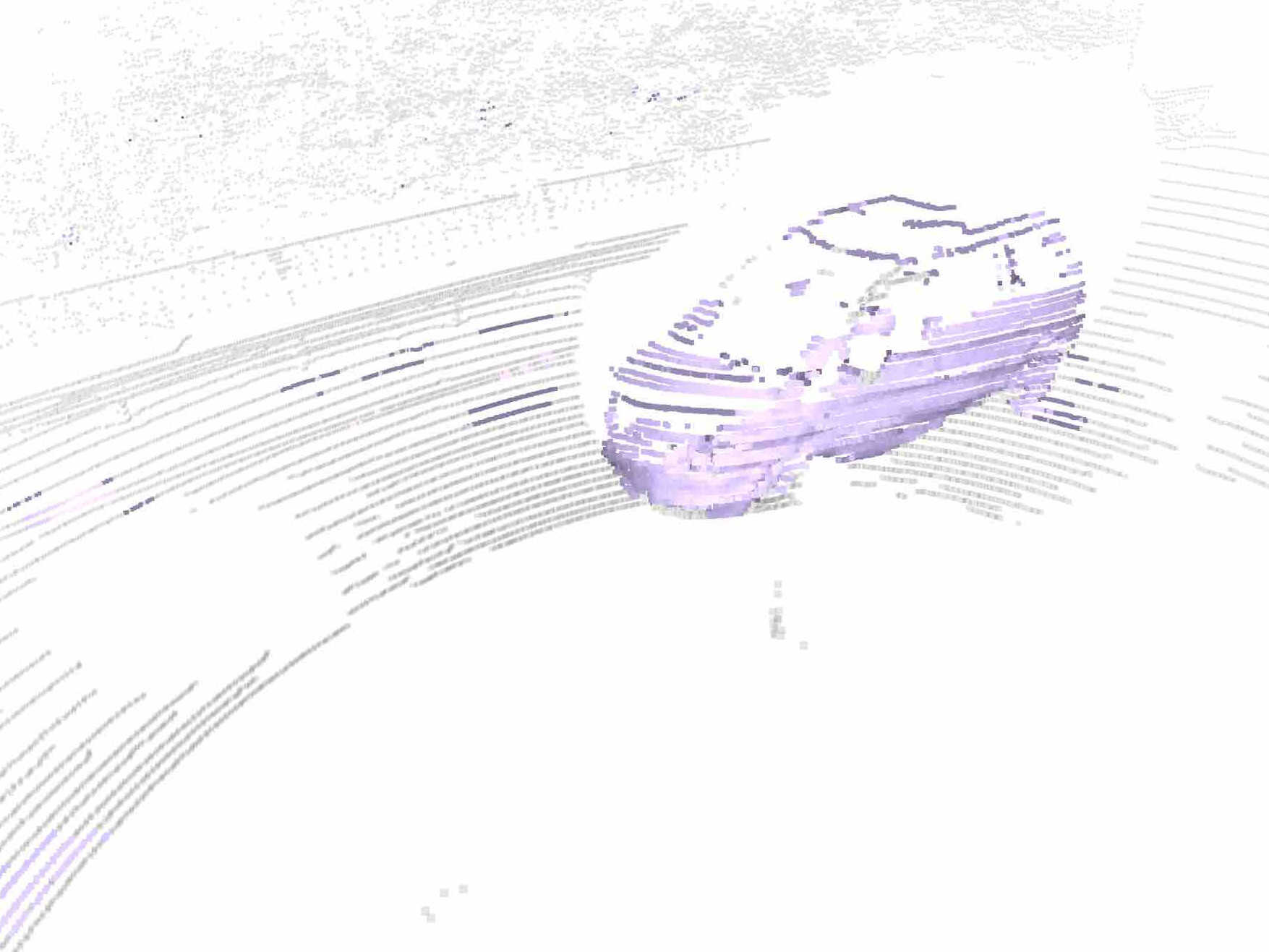}\put(300,100){\makebox(0,-10){\color{black}\textbf{\scriptsize IOU=76.6}}\color{black}}
\end{overpic} \\
\end{tabular}
\caption{\textbf{Qualitative results} with clicks and corresponding segmentation masks. Our approach accurately segments objects from different datasets with as little as \textbf{one user click} (\textit{top row, right}). ScanNet*~(\textit{first row}) shows classes that have not been used in training: \textit{television (left)} and \textit{towel (right)}.\\}
\label{fig:qual}
\vspace{-10pt}
\end{figure*}

We show visualizations on full scenes in ~Fig.~\ref{fig:teaser} and of object segmentation on all three datasets in Fig.~\ref{fig:qual}.
Our method predicts accurate segmentations in heavily cluttered environments with very limited user clicks. %

\subsection{User Study}
To evaluate our method with real human annotators,
we also conducted a small-scale experiment replacing the simulated annotator with three real human annotators in the loop.
To that end, we implemented an interactive annotation tool that runs our model at the backend.
The annotators engaged in the user study had no prior experience with annotating.
We explained them how the tool works and allowed them to label a few examples to familiarize themselves with the tool before recording their behavior.
The annotators see the full scene as does the simulated annotator without cropping or using bounding boxes.
If they deemed the segmentation satisfactory, they could submit the result as final at anytime. 

\begin{table}[b]
\caption{\textbf{Real Human Annotators User Study.}  Human annotators labeled points of the class ``picture'' and of random objects showing that our simulation strategy is representative of real human input and produces similar results.}
\vspace{-5px}
\centering
\begin{minipage}{0.67\linewidth}
\centering
\resizebox{0.99\linewidth}{!}{%
\setlength{\tabcolsep}{3pt}
	\begin{tabular}{lrc|cc}
	\toprule
	 & \multicolumn{2}{c|}{\textbf{ScanNetV2}-\textit{picture}}& \multicolumn{2}{c}{\textbf{ScanNetV2}-\textit{random}}\\
     &  \multicolumn{1}{c}{\textbf{\iou{}@5}} & \multicolumn{1}{c|}{\textbf{\iou{}@10}} &\multicolumn{1}{c}{\textbf{\iou{}@5}} & \multicolumn{1}{c}{\textbf{\iou{}@10}} \\
	\arrayrulecolor{lightgray}\hline\arrayrulecolor{black}
	\arrayrulecolor{lightgray}\hline\arrayrulecolor{black}
	Human  &  49.6  & 57.0 & 78.1 & 82.0  \\
	Simulator  &  32.0  & 54.5 & 79.8 &  84.8\\
    \bottomrule
	\end{tabular}
	}
	\end{minipage}
	\raisebox{-0.cm}{
    \begin{minipage}{0.29\linewidth}
        \centering
        \includegraphics[width=1\linewidth, trim={0 70px 0 0}, clip]{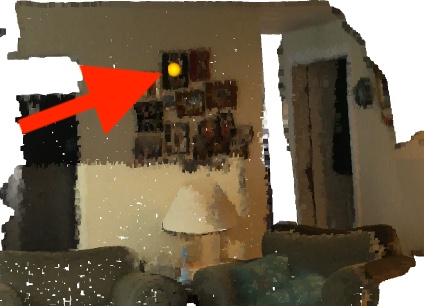}
	\end{minipage}
	}
 \label{tab:user_study}

\end{table}

First, we asked the annotators to label 10 random objects from the challenging class \textit{picture} using our interactive segmentation tool and we compared the real users results with the corresponding results from the simulated annotator~(\textit{Task~1}).
Then we asked the annotators to randomly segment 20 objects from ScanNetV2 val. set and compared them with the same objects annotated by the simulated annotator~(\textit{Task~2}). 
The results are presented in Tab.~\ref{tab:user_study}.
Task~1 is especially challenging as it can be seen from the example figure of our user interface (right). The geometry
of the points belonging to the class is very similar to the one of the wall
nearby and only color can help distinguish the separate objects. Despite that,
real human annotators performed actually slightly better than the simulated
clicks especially in the low clicks regime. Labeling random objects also performs similarly to the simulated clicks. 

We note that although our model is trained with simulated clicks, it shows robustness to human input even though they follow different labeling strategies and that our simulation of interactive object segmentation~(Sec.~\ref{sec:method}) is representative of real human input and both provide similar output.

\subsection{Inference Speed.} 
After interaction, the user sees the updated segmentation mask in~\textbf{0.06 sec} (single NVIDIA~Titan~RTX), making the method suitable for real-time point cloud annotation.\\

\section{Conclusion}
In this work, we introduced, for the first time, a method for interactive object segmentation on 3D points clouds.
Our approach is a simple yet powerful framework that achieves highly accurate 3D segmentation masks with little human annotation effort.
Our method generalizes well to different datasets that have not been trained on and enables the annotations of novel large-scale 3D training sets including previously unseen semantic classes.
In particular, we also presented practical policies necessary for simulating user clicks during training and evaluation, verified by a user study.

\balance

\newpage
\vspace{5px}
\bibliographystyle{plain}
\balance
\bibliography{longstrings,interactive3d,loco,3d}
\end{document}